\documentclass[twocolumn]{svjour3}
\usepackage[utf8]{inputenc}
\usepackage{epsfig} % for postscript graphics files
\usepackage{amsfonts,latexsym,amssymb} %paquets mathï¿œmatiques
\usepackage{mathrsfs,amsmath}
\usepackage{graphicx}
\usepackage{xcolor}
\usepackage{hyperref}
\usepackage{caption}
\usepackage{subcaption}
\usepackage{tikz}

\usepackage{mathptmx}
\usetikzlibrary{arrows,matrix}
\usepackage{epstopdf}
\usepackage{multirow}
\usepackage{array}
\usepackage{hyperref}
\usepackage{amsmath}

\DeclareMathAlphabet      {\mathbfit}{OML}{cmm}{b}{it}

\def \begpf{ \begin{description} \item[{\bf Proof:\hspace{1.5em} }]  }
\def \endpf{ \begin{flushright} $\Box $ \end{flushright}   \end{description}}

    \newtheorem{prop}{Proposition}[section]
    \newtheorem{lemm}[prop]{Lemma}

    \newtheorem{exmpl}{Example}
    
    \newcommand{\begex}[2]{ \begin{exmpl}[#1] \label{#2} \end{exmpl}
    \begin{description}  \item
   }
    \def  \endex {       \end{description}  }

\def \begpf{ \begin{description} \item[{\bf Proof:\hspace{0.75em} }]  }
\def \endpf{ \begin{flushright} $\Box $ \end{flushright}   \end{description}}
\def \begrm{ \begin{description} \item[{\bf Remark:\hspace{0.5em} }]  }

\def \endrm{  \end{description} }
\def \begex{ \begin{description} \item[{\bf Example:\hspace{0.5em} }]  }
\def \endex{  \end{description} }

\newcommand{\beq}[1]{ \begin{equation} \label{eq.#1} }
\newcommand{\eeq}{ \end{equation} }

\newcommand{\blemm}[1]{ \begin{lemm} \label{lemm.#1} }
\newcommand{\elemm}{ \end{lemm} }
\newcommand{\barr}{ \begin{array}} %{cl} }
\newcommand{\earr}{ \end{array} }

\begin{document}
\title{\centering Dynamics and multi-stability of a rotor-actuated Twistcar robot\linebreak with passive steering joint}
\author{Anna Zigelman$^{1,\ast}$\thanks{$^{\star}$Corresponding Author \email{annar@technion.ac.il}} \and Zitao Yu$^{1,2,\dagger}$\thanks{$\dagger$The work of Z. Yu on this research has been conducted as part of his MSc  Thesis at Technion.} \and Rom Levy$^{1}$ \and  Yizhar Or$^{1}$}

%\footnote{}\thanks{$^{\star}$Corresponding Author \email{annar@technion.ac.il}}
%\institute{%\authorrunning{Short form of author list} % if too long for running head
%%E. Tovi \at
%              A. Z., Z. Y., R. L., and Y. O. are with the Faculty of  Mechanical Engineering,
%Technion - Israel Institute of  Technology, Haifa 3200003, Israel
%}

\institute{%\authorrunning{Short form of author list} % if too long for running head
	%E. Tovi \at
	1. Faculty of  Mechanical Engineering,
	Technion - Israel Institute of  Technology, Haifa 3200003, Israel.\\
	2. Deptartment of Aerospace and Mechanical Engineering, University of Southern California, USA.
}

\maketitle

\begin{abstract}
The nonlinear dynamics of many under-actuated wheeled platforms are governed by nonholonomic constraints of no-skid for passively rolling wheels, coupled with momentum balance. In most of theoretical models, the shape variables, i.e. joint angles, are directly prescribed as periodic inputs, such as steering angle of the Twistcar. In this work, we study a variant of the Twistcar model where the actuation input is periodic oscillations of an inertial rotor attached to the main body, while the steering joint is passively free to rotate. Remarkably, the dynamics of this model is extremely rich, and includes multiplicity of periodic solutions, both symmetric and asymmetric, as well as stability transitions and bifurcations. We conduct numerical simulations as well as asymptotic analysis of the vehicle’s reduced equations of motion. We use perturbation expansion in order to obtain leading-order dynamics under symmetric periodic solution. Then, we utilize harmonic balance and further scaling assumptions in order to approximate the conditions for symmetry-breaking pitchfork bifurcation and stability transition of the symmetric periodic solution, as a function of actuation frequency and structural parameters. The asymptotic results show good agreement with numerical simulations. The results highlight the role of passive shape variables in generating multi-stable periodic solutions for nonholonomic systems of robotic locomotion. 

\end{abstract}


%\section*{Declarations}
%Funding: This research has been supported by Israeli Science Foundation (ISF) under grant no. 1005/19, and by the Israeli Ministry of Science and Technology under grant no. 3-15622.
%
%Conflicts of interest/Competing interests: The authors have no conflict of interest.
%
%Availability of data and material: Not applicable.
%
%Code availability: Not applicable.

\section{Introduction} 
Many under-actuated wheeled platforms contain axles of passively rolling wheels. Examples are various skating boards such as snakeboard and carver board, as well as children’s toy cars such as Twistcar. The dynamic motion of such platform is governed by enforcing zero lateral slip (i.e. skid) of the wheels, coupled with actuation of periodic inputs. No-skid constraints are nonholonomic in their nature~\cite{Bloch,Neimark}, constraining the system’s instantaneous allowable velocity directions, similar to ice-skating blades. 

Early mathematical examples of nonholonomic systems included planar kinematic models of a car with controlled steering (Dubins’ car~\cite{Macharet}), or truck connected to a chain of passive wheeled trailers via hinged passive joints~\cite{nakamura2001design}. The inputs to such systems were kinematic – driving speed and prescribed steering angle or its angular velocities. The main challenges in such systems were planning a motion trajectory that respects all nonholonomic constraints~\cite{pardi2020path}, and feedback stabilization of tracking the desired path~\cite{yang2003tracking}. Another example is the three-wheel snake robot~\cite{Shammas_2007,Yona_2019}, which has three axles of passive wheels and two joint angles prescribed as periodic inputs (gaits). Such systems were analyzed utilizing concepts of geometric mechanics~\cite{Kelly_1995,Ostrowski_1998}. Other related models of three-link two-joint systems governed by first-order constraints on shape- and body-velocities are Purcell’s swimmer in highly viscous flow~\cite{becker2003self}, and inertial three-link swimmer in ideal fluid~\cite{Kanso_2005}, which were analyzed using similar geometric tools~\cite{Gutman_2016,Hatton_2013}.
 
In cases where the number of nonholonomic constraints is less than the number of passive degrees-of-freedom, the system is no longer “kinematic” and its motion is governed by coupling of momentum balance, nonholonomic \linebreak constraints, and the periodic actuation. An early basic example of such dynamic-nonholonomic system is Chaplygin’s sleigh~\cite{chaplygin_collected_nonholonomic_1950,STANCHENKO198911}, a planar passive rigid body with a point blade, which is allowed to move only in a specific body-fixed direction. Later works studied variants of Chaplygin’s sleigh with different types of added oscillatory actuation~\cite{Kelly,Osborne,Tallapragada}, and also with incorporation of frictional energy dissipation~\cite{Bizyaev,Fedonyuk_2017,fedonyuk2018sinusoidal}. Several theoretical works have studied skating and riding toys in the framework of underactuated robots, including snakeboard~\cite{Ostrowski_1994}, tricycle~\cite{chitta2005robotrikke} and several variants of the roller-racer~\cite{Bizyaev_2018,Krishnaprasad}, which was also named Landshark~\cite{Bazzi} and Twistcar~\cite{Chakon,Halvani}. Some of these works utilized concepts of geometric mechanics for analyzing the robot’s net motion under oscillatory inputs~\cite{Ostrowski_1998,Ostrowski_1995}, whereas other works employed asymptotic analysis~\cite{Chakon,Halvani} and other methods~\cite{artemova2024dynamics,Bazzi}. Few of these works also included experiments with robotic prototypes~\cite{chitta2005robotrikke,Kilin,Levy,salman2016physical}. The periodic inputs of such robots and models commonly involve relative angles at joints connecting the links, such as two joint angles of the three-wheel snake~\cite{rizyaev2024locomotion,Shammas_2007,Yona_2019} and a single steering joint angle of the roller-racer/Twistcar/Landshark \cite{Bazzi,Chakon,Krishnaprasad}. Other options are an oscillating mass on top of the main body of Chaplygin’s sleigh~\cite{Osborne}, and angular oscillations of an inertial rotor as in the “Chaplygin’s Beanie” model~\cite{Kelly}, or in the snakeboard~\cite{Ostrowski_1994}, whose rotor represents waist twisting motion of a human rider standing on the board. 

In the models mentioned above, all shape variables, i.e. joint angles, are directly prescribed as inputs. In contrast, recent works have studied underactuated passive-wheeled multi-link robots such that at least one internal joint is passive.  In most cases, these passive joints are acted by torsional spring and damper, representing structural visco-elasticity \cite{Dear_2020,Dear_2016}. This adds dynamic coupling of the internal shape variable with the body motion and periodic actuation, which often results in existence of optimum mean speed due to its non-monotonic dependence on frequency of the oscillatory input~\cite{rizyaev2024locomotion}. 

Importantly, recent theoretical works also found that incorporating passive visco-elastic shape variables may also introduce multiplicity of periodic motions, as well as their stability transitions and bifurcations. While these effects have been studied for multi-link swimmer models with internal \cite{TOVI2024,Zigelman} or external actuation \cite{Harduf,Paul}, they were recently found also in the work~\cite{Rodwell}, which considered a simple modification of Chaplygin sleigh model. The planar two-link model in~\cite{Rodwell} consists of a sleigh with a single no-skid blade, connected to a second link by a revolute joint with torsional spring and viscous damper, in parallel to periodic input of joint torque actuation. The work showed that one can tune the spring’s potential elastic energy in order to induce existence of asymmetric fixed points of joint angle, leading to multiplicity of periodic solutions oscillating about those fixed points. Furthermore, these limit cycles may undergo stability transitions upon varying the input actuation frequency. The interesting results in~\cite{Rodwell} call for extension and exploration of other models of dynamic underactuated nonholonomic systems with passive shape variables, as well as deeper investigation using asymptotic analysis. 

\begin{figure}%[h]
	\centering
	\includegraphics[width=0.5\textwidth]{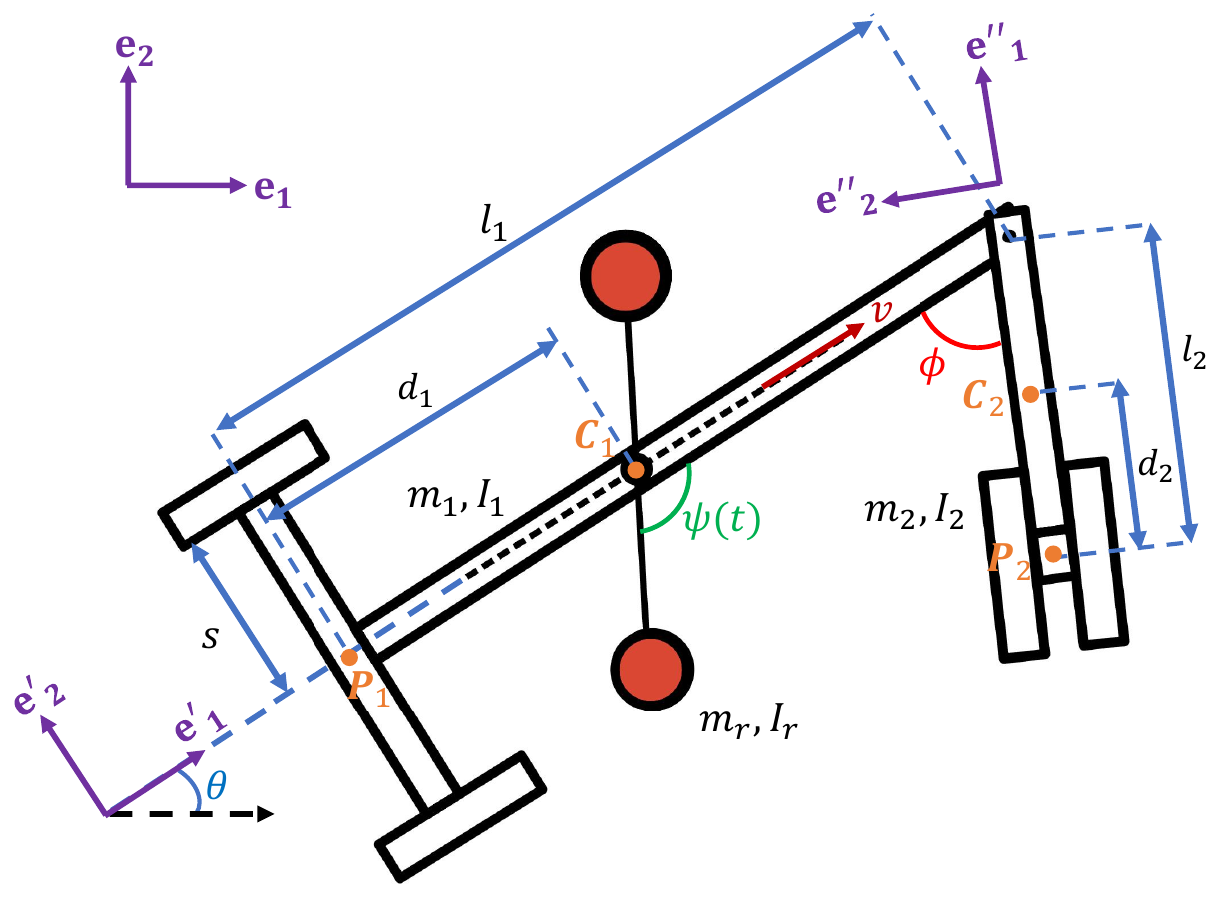}
	\caption{The RAPS-Twistcar model.} \label{fig1}
\end{figure}

In our current work, we study a variant of the planar Twistcar model in~\cite{Chakon}, called RAPS Twistcar \linebreak (Rotor-Actuated Passive Steering Twistcar). The model, introduced in Section~\ref{Sec2} below, consists of two links supported by two no-skid wheeled axles with rolling resistance of viscous dissipation, see Fig.~\ref{fig1}. Inspired by the snakeboard~\cite{Ostrowski_1994}, the actuation input is periodic oscillations of an inertial rotor attached to the main body, while the steering joint is passively free to rotate, with zero torque and no elasticity or friction. Remarkably, numerical simulations (Section~\ref{Sec3}) reveal that the dynamics of this model is extremely rich, and includes multiplicity of periodic solutions, both symmetric and asymmetric, as well as stability transitions and bifurcations. After simplifying reduction of the system's dynamic equations of motion in Section~\ref{S:Reduction of the system}, we conduct in Section~\ref{S:Analysis of periodic solutions} numerical Poincaré map analysis~\cite{Guckenheimer} for numerically finding periodic solutions and examine their stability. This reveals bifurcations, stability transitions and \linebreak symmetry-breaking of periodic solution branches. Next, in Section~\ref{S:4} we present asymptotic analysis of the vehicle’s reduced equations of motion assuming small-amplitude oscillations of the input. We use perturbation expansion~\cite{Nayfeh_2008} to obtain the system’s approximate leading-order dynamics under symmetric periodic solution, which enables finding optimal placement of the bod’s center of mass. Finally, in Section~\ref{S:44} we utilize harmonic balance~\cite{Nayfeh_Mook_2008} and further scaling assumptions in order to approximate the conditions for symmetry-breaking pitchfork bifurcation and stability transition of the symmetric periodic solution, as a function of actuation frequency and structural parameters. The asymptotic results show good agreement with numerical simulations. Our results highlight the significant role of passive shape variables in generating multi-stable periodic solutions for nonholonomic systems of robotic locomotion, and the importance of asymptotic analysis for improved understanding of these intriguing phenomena.

\section{Problem formulation}\label{Sec2}
%The physical simplified model is illustrated in Fig.~\ref{fig1}, which is based on the twistcar model. It is composed of two rigid links: the longer one represents the rider and the twistcar body, while the shorter one represents the lever of the front wheels. The mass and inertia of the main body link are $m_1$, $I_1$, the body mass and inertia of the rider are $m_2$, $I_2$, and the mass and inertia of the frontal link are $m_3$, $I_3$. Since this model is directly developed from
%the dissipative twistcar model, nonholonomic constraints exist as the no-side-skidding conditions on the wheels, and the dissipation is assumed for the rolling resistance of the wheels. Formulations aiming at obtaining the reduced equation of motion are developed in the same
%structure as in~\cite{} (ADD REFERENCE), in the dissipative twistcar and skidding twistcar model.

The simplified physical model of the RAPS-Twistcar is shown in Fig.~\ref{fig1}. Like the Twistcar model~\cite{Chakon}, this system consists of two rigid links supported by wheeled axles: the longer link represents the rider and the body of the Twistcar, while the shorter link represents the steering lever controlling the front wheels. We denote the mass and inertia of the longer link by $m_1$ and $I_1$, the body mass and inertia of the rider by $m_r$, $I_r$, and the mass and inertia of the frontal link by $m_2$, $I_2$.
% and assume that $m_1$ and $I_1$ are significantly larger than $m_2$ and $I_2$. 
We assume that the front and rear wheels cannot skid sideways (no-skid conditions), which are introduced as nonholonomic constraints. 
%Moreover, we assume that the shape variable of the system is actuated with a simple periodic time-series input $\varepsilon \cos(\omega t)$.
% + \phi_0$. In this work, we define the Twistcar configuration with $\phi_0 = 0$.

The system's motion is described by $\mathbf{q}$, consisting of five generalized coordinates, 
\begin{equation} \label{generalized coordinates}
	\mathbf{q}=( x,y,\theta, \phi,\psi)^T,
\end{equation}
where $(x,y)$ denote the position of the point $P$ in world-fixed frame $\{\mathbf{e}_1,\mathbf{e}_2\}$, $\theta$ is the orientation angle of the body link of the car with respect to world frame, $\phi$ is the relative angle between the two links, and $\psi$ is the relative angle between the rotor and the longer link. The rotor angle $\psi$ is used as the input to the system,
\begin{equation} \label{Psi Input}
	\psi(t) = A \sin{(\Omega t)},
\end{equation}
where $A$ is the amplitude, $t$ is time, and $\Omega$ is frequency. We define the $x,y$ and $\theta$ as the body coordinates and $\phi$ as the shape variable.
%, which describes the locomotion of the whole robot in the world frame and the configuration status of the robot correspondingly. 
%In this problem, prescribe $\phi(t) = \varepsilon \cos(\Omega t) + \phi_0$. 

The kinetic energy in this system may be expressed as,
\begin{equation} \label{Lagrangian - primary}
	\begin{aligned}
		E_K&= \frac{1}{2} (m_1+m_r) (\mathbf{v}_{C_1} \cdot \mathbf{v}_{C_1}) + \frac{1}{2} m_2 (\mathbf{v}_{C_2}\cdot\mathbf{v}_{C_2})\\
		&+ \frac{1}{2}I_1\dot{\theta}^2 + \frac{1}{2}I_2(\dot\theta+\dot\phi)^2+\frac{1}{2}I_r(\dot{\psi}+\dot{\theta})^2,
	\end{aligned}
\end{equation}
where $\textbf{v}_{C_i}$ is the velocity of point $\mathbf{C}_i$, which is the center-of-mass (COM) of link $i$ for $i=1,2$ (see Fig.~\ref{fig1}).

The nonholonomic no-skid constraints can be expressed as $\mathbf{v}_{P_1}\cdot\mathbf{e}_2'=0$ and $\mathbf{v}_{P_2}\cdot\mathbf{e}_2''=0$, where $\{\mathbf{e}_1',\mathbf{e}_2'\}$ frame is attached to the body link, $\{\mathbf{e}_1'',\mathbf{e}_2''\}$ frame is attached to the frontal link, and $\textbf{v}_{P_i}$ is the velocity of point $\mathbf{P}_i$ for $i=1,2$ (see Fig.~\ref{fig1}). These constraints may be expressed in a matrix form, as
\begin{equation} \label{world frame constraint equations}
	\begin{aligned}
		& \mathbf{W}(\mathbf{q}) \dot{\mathbf{q}} =0,\\
		& \mathbf{W}(\mathbf{q}) = 
		\begin{pmatrix}
			-\sin{(\theta)}    &    \cos{(\theta)}    &   0  & 0 & 0\\
			-\sin{(\phi + \theta)} &  \cos{(\phi + \theta)} &  l_1\cos{(\phi)}-l_2 & -l_2 &0
		\end{pmatrix}.
	\end{aligned}
\end{equation}

In order to add viscous rolling resistance to the model, we use Rayleigh's dissipation function, as follows. For each wheel, we define the velocity along its permissible roll direction using the Jacobian,
\begin{equation} \label{Roll Velocity - with Jacobian}
	\textbf{v}_{\parallel, i} = \mathbb{J}_i(\textbf{q}) \dot{\textbf{q}}.
\end{equation}

Using the Jacobian notation, we define Rayleigh's dissipation function, 
\begin{equation} \label{Rayleigh Dissipation Function}
	\mathcal{R}(\textbf{q}, \dot{\textbf{q}}) = \frac{c}{2} \sum_{i=1}^{3}{ {\textbf{v}_{\parallel, i}}^{T} \textbf{v}_{\parallel, i}} = \frac{c}{2} {\dot{\textbf{q}}}^T \left( \sum_{i=1}^{3}{{\mathbb{J}_i(\textbf{q})}^T \mathbb{J}_i(\textbf{q})} \right) \dot{\textbf{q}},
\end{equation}
where \(c\) is the viscous damping coefficient.

We differentiate \eqref{Rayleigh Dissipation Function} with respect to the generalized velocity \(\dot{\textbf{q}}\) to obtain the dissipation contribution to the dynamic equations. Using the Jacobian notation, we obtain that
\begin{equation} \label{Dissipation Vector}
	\textbf{D}(\textbf{q}, \dot{\textbf{q}}) = \frac{\partial \mathcal{R}}{\partial \dot{\textbf{q}}} = c \left( \sum_{i=1}^{3}{{\mathbb{J}_i(\textbf{q})}^T \mathbb{J}_i(\textbf{q})} \right) \dot{\textbf{q}}.
\end{equation}
 
%Moreover, note that a planar system without elastic components has $E_P=0$, such that the Lagrangian in this system may be expressed as,
%\begin{equation} \label{Lagrangian - primary}
%	\begin{aligned}
%	\mathbf{L}&=E_K\\
%	&= \frac{1}{2} (m_1+m_r) (\mathbf{v}_C \cdot \mathbf{v}_C) + \frac{1}{2} m_2 (\mathbf{v}_Q\cdot\mathbf{v}_Q)\\
%	&+ \frac{1}{2}I_1\dot{\theta}^2 + \frac{1}{2}I_2(\dot\theta+\dot\phi)^2+\frac{1}{2}I_r(\dot{\psi}+\dot{\theta})^2,
%	\end{aligned}
%\end{equation}
%where $\mathbf{v}_C$ and $\mathbf{v}_Q$ are the velocities at points $C$ and $Q$, respectively (see Fig.~\ref{fig1}).
%Next, we use the Rayleigh dissipation function to formulate the assumed viscous friction, which is given by
%\begin{equation} \label{Dissipative Twistcar Rayleigh Dissipative function}
%	D=\frac{1}{2} c_w (\mathbf{v}_{RL}\cdot\mathbf{e}_1')^2 + \frac{1}{2} c_w (\mathbf{v}_{RR}\cdot\mathbf{e}_1')^2 + \frac{1}{2} c_w (\mathbf{v}_{F}\cdot\mathbf{e}_1'')^2,
%\end{equation}
%where $\mathbf{v}_{RR}$ and $\mathbf{v}_{RL}$ are the velocities at points $RR$ and $RL$, respectively (see Fig.~\ref{fig1}). 

Next, following~\cite{Levy}, we derive the equations-of-motion of the system. More specifically, using the Euler-Lagrange formulation \cite{Murray_1994}, with the assumption of a planar model with a constant potential energy, we obtain:
%Since the system is planar, without elastic components so that $E_P=0$, Next, let us employ the equation of motion, 
\begin{equation} \label{organized constraints Lagrange Equation}
	\mathbf{M}(\mathbf{q}) \mathbf{\Ddot{q}} + \mathbf{B}(\mathbf{q},\mathbf{\dot{q}}) + \mathbf{D}(\mathbf{q},\mathbf{\dot{q}})= \mathbf{F}_q + \mathbf{W}^T(\mathbf{q}) \mathrm{\pmb{\upLambda}},
\end{equation}
where $\mathbf{M}(\mathbf{q})$ is the system's matrix of inertia, $\mathbf{B}(\mathbf{q},\mathbf{\dot{q}})$ is the vector of velocity-dependent inertial forces and $\mathbf{D}(\mathbf{q},\mathbf{\dot{q}})$ is the vectors of dissipation forces.  $\pmb{\upLambda}= \begin{bmatrix} \lambda_1 & \lambda_2 \end{bmatrix}^T$ is the vector of forces enforcing the constraints and $\textbf{F}_q = \begin{bmatrix}0 & 0 & 0 & 0 & \tau\end{bmatrix}^T$ is the vector of generalized forces, which includes the applying an outside torque $\tau$ on the rotor angle.

%where $\pmb{\Lambda}= \begin{bmatrix} \lambda_1 & \lambda_2 \end{bmatrix}^T$ is the vector of forces enforcing the constraints, $\mathbf{F}_d =dD/d\dot{\mathbf{q}}$,  and $\mathbf{W}$ is given in~\eqref{world frame constraint equations}. For the explicit expressions for $\mathbf{M}$, $\mathbf{B}$, and $\mathbf{F}_d$ see equations~\eqref{E:M and B} and~\eqref{explicit expression for the dissipation force} in Appendix.

By differentiating \eqref{world frame constraint equations} with respect to time, we get two more equations, which may be expressed as
\begin{equation} \label{Constraints Equation - matrix form}
	\textbf{W}(\textbf{q}) \ddot{\textbf{q}} + \dot{\textbf{W}}(\textbf{q}) \dot{\textbf{q}} = 0.
\end{equation}

\section{Numerical simulations}\label{Sec3}
The system of equations in~\eqref{organized constraints Lagrange Equation}-\eqref{Constraints Equation - matrix form} is differential-algebraic, which is possible to solve by numeric integration, for example by using ``ode45'' solver in Matlab. To simulate this system we use the parametric values in Table~\ref{Tab:1}, where note that we neglect the mass and inertia of the vehicle's links with respect to the rotor, so that $m_1$ and $m_2$ are negligible with respect to $m_r$. Similarly, $I_1$ and $I_2$ are negligible with respect to $I_r$. Hence, hereafter we assume that $m_1=m_2=I_1=I_2=0$. Note that although this assumption implies that the matrix $\mathbf{M}$ is singular, the system of equations in~\eqref{organized constraints Lagrange Equation}-\eqref{Constraints Equation - matrix form} is non-singular, where $\psi(t)$ and its derivatives are known.
%Moreover, we assume that joint dissipation vanishes. 

%In the following simulations, we use the parameter values as listed in Table~\ref{Tab:1}:
\begin{table}[!h]
	\centering
	%	\fontsize{8}{12pt}\selectfont
	\begin{tabular}{ |m{1.1cm}|l|m{3.45cm}| }
		\hline
		Parameter   & Value & Description   \\
		\hline
		$l_1$  &  0.6 [m] & main body length\\
		\hline
		$l_2$  &  0.2 [m] & frontal link length \\
		\hline
		$d_1$ & 0.06 [m]& main body COM location\\ 
		\hline
		$d_2$ & 0.1 [m] & frontal link COM location\\
		\hline
		$s$ & 0.2 [m] & rear wheels width\\
		\hline
		$m_r$ & 40 [kg] &rotor mass\\
		\hline
		$I_r$ & 0.1695 [kg$\cdot$m$^2$] & rider inertia\\
		\hline
		$c$ & 10 [N$\cdot$s/m] & rolling dissipation coefficient\\
		\hline
		$A$& 1 [rad] & actuation amplitude\\
		\hline
	\end{tabular}
	\caption{Dimensions and values of the model's physical parameters.}
	\label{Tab:1}
\end{table}

\begin{figure*}[h!]
	\centering
	\includegraphics[width=\textwidth]{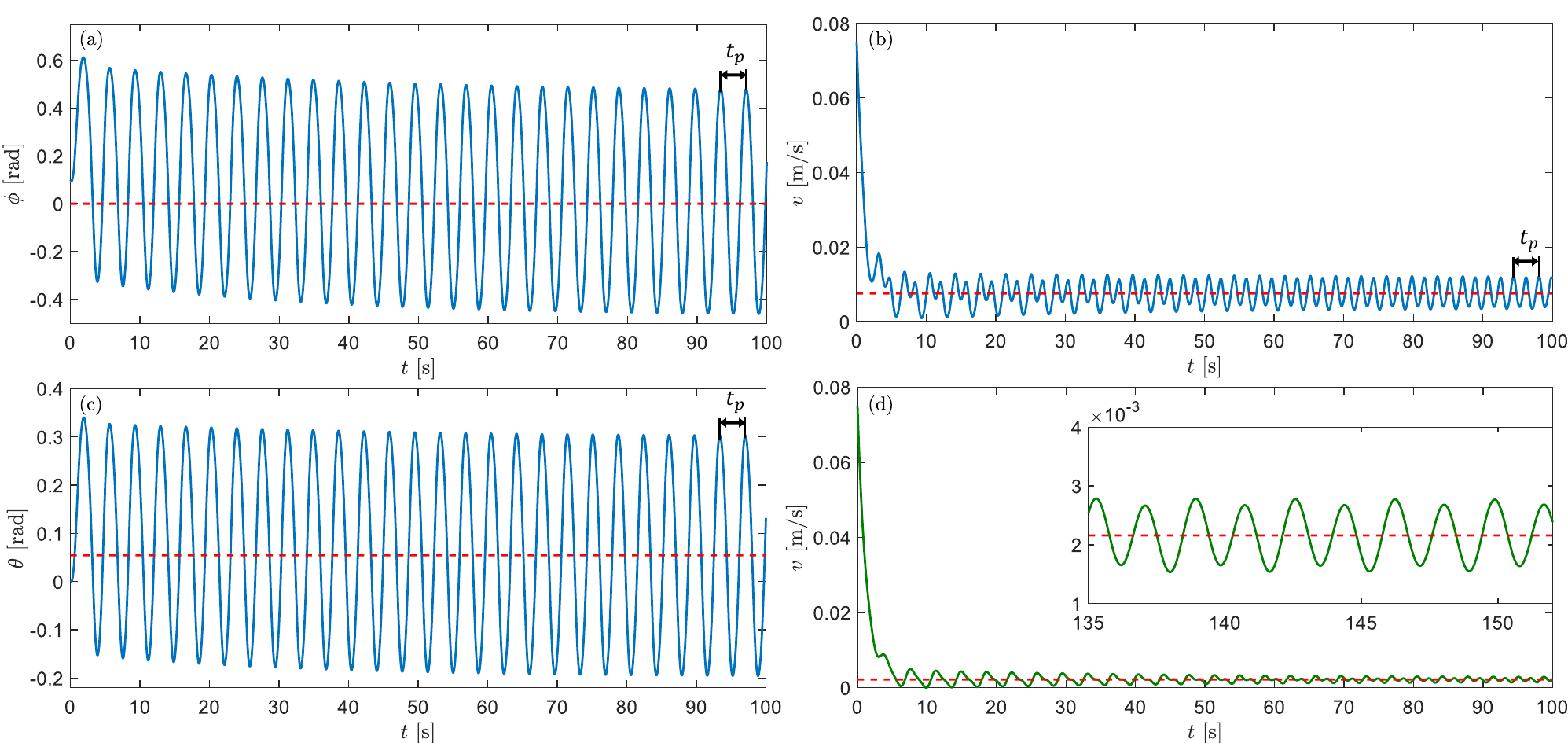}
	\caption{Numerical simulation for ``high'' frequency $\Omega=1.72$ [rad/s] resulting in convergence to a symmetric periodic solution, where the physical parameters are given in Table~\ref{Tab:1}. (a) The passive steering angle $\phi$ versus time $t$. (b) The speed $v$ in direction parallel to the main link versus time $t$. (c) The body orientation angle $\theta$ versus time $t$. (d) The speed $v$ in direction parallel to the main link versus time $t$, for the parameter values as given in Table~\ref{Tab:1}, except that now $d_1=0.12$ [m]. In the inset of (d) we show magnified view of the convergence to the steady-state. The steady-state mean values $\bar{\phi}$, $\bar{v}$, $\bar{\theta}$, and $\bar{v}$ are indicated by red dashed lines in panels (a)-(d), respectively.} \label{fig2A}
\end{figure*}

We hereby present results of various numerical simulations. In Fig.~\ref{fig2A}(a) we show the dependence of the passive joint angle $\phi$ on time $t$. We used the parameters from Table~\ref{Tab:1} and assumed that the actuation of the simulation is $\psi(t)=A\sin{(\Omega t)}$, with the amplitude of $A=1$ [rad] and frequency $\Omega=1.72$ [rad/s]. The solution is composed of an initial transient which converges to a steady-state periodic solution. It is possible to see that $\phi(t)$ converges to a periodic signal with frequency which equals to the actuation frequency. 

\begin{figure*}[h!]
	\centering
	\includegraphics[width=\textwidth]{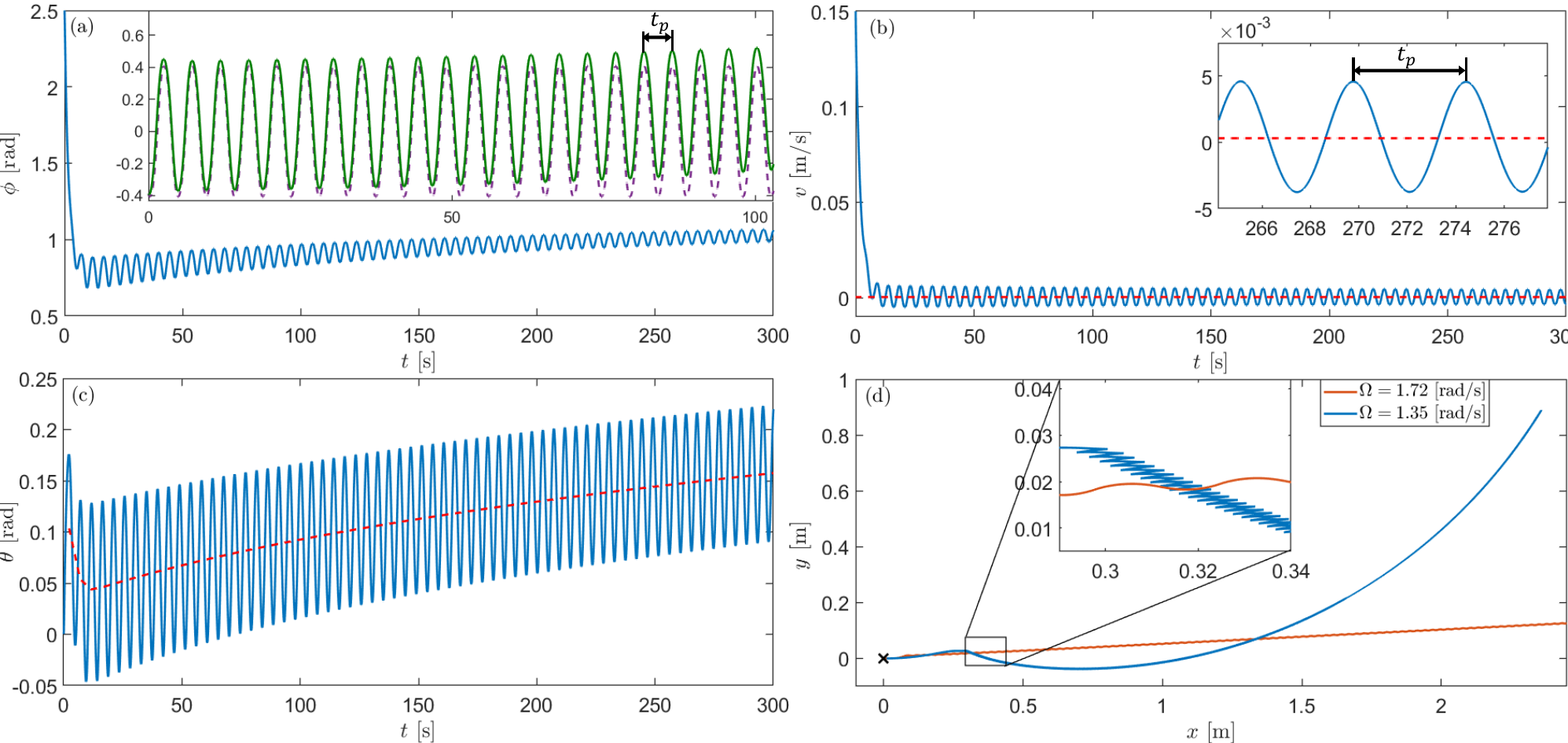}
	\caption{Numerical simulation for ``low'' frequency $\Omega=1.35$ [rad/s], where the physical parameters are given in Table~\ref{Tab:1}. (a) The passive steering angle $\phi$ versus time $t$, which converges to asymmetric periodic solution, where in the inset we show an unstable symmetric solution with `exact' initial conditions (purple dashed curve) and with slightly perturbed initial conditions (green curve). (b) The speed $v$ in direction parallel to the main link versus time $t$, where its steady-state mean value $\bar{v}$ is indicated by a red dashed line. In the inset we show a magnified view of (b) near the final time of the simulation. (c) The body orientation angle $\theta$ versus time $t$, where its mean value $\bar{\theta}$ over a period is indicated by a red dashed line. (d) Comparison between the trajectories in the $xy-$plane in the asymmetric case (blue curve) with the frequency of $\Omega=1.35$ [rad/s] (as in the rest of the panels in this figure) and the symmetric case (orange curve) with the frequency of $\Omega=1.72$ [rad/s] (as in Fig.~\ref{fig2A}), where in the inset we show a magnified view of (d) in the vicinity of the initial position, which is marked with `x'.} \label{fig3A}
\end{figure*}

Let us define the mean value $\bar{\phi}$ about which the angle $\phi(t)$ oscillates as the mean over a period, after initial
transient response followed by convergence to the periodic solution for $\phi$. Similarly, we define the mean values $\bar{\theta}$ and $\bar{v}$, namely
\begin{equation}\label{E:def means}
	\begin{aligned}
	&\bar{\phi}=\frac{1}{t_p}\int_{t_0}^{t_0+t_p}\phi(t)dt, \quad \bar{\theta}=\frac{1}{t_p}\int_{t_0}^{t_0+t_p}\theta(t)dt,\\
	&\bar{v}=\frac{1}{t_p}\int_{t_0}^{t_0+t_p}v(t)dt,
	\end{aligned}
\end{equation}
where $t_p=2\pi/\Omega$ denotes the period of the actuation $\psi(t)$ and $t_0$ is an arbitrary time after reaching the steady-state behavior. For the simulation shown in Fig.~\ref{fig2A}(a), $\bar{\phi}=0$ is indicated by red dashed line. This means that the steady-state periodic solution is \emph{symmetric} relative to the main body link.

We define the velocities $v_{\parallel}$ and $v_{\perp}$ as the velocity components of point $P_1$ along $\textbf{e}_1'$ and $\textbf{e}_2'$ directions, respectively, and for brevity denote $v=v_{\parallel}$. In Fig.~\ref{fig2A}(b), we show the speed $v$ of the twist-car robot versus time $t$. It can be seen that after an initial transient the frequency of the periodic function $v(t)$ becomes doubled relative to the frequency of the actuation $\psi(t)$. Moreover, the steady-state function $v(t)$ oscillates about a constant $\bar{v}\neq 0$, which is indicated by red dashed line in Fig.~\ref{fig2A}(b). In Fig.~\ref{fig2A}(c), we show the passive angle $\theta$ versus time $t$. It can be observed that after the initial transient the passive angle $\theta(t)$ oscillates about a constant value $\bar{\theta}\neq 0$. In Fig.~\ref{fig2A}(d), we show the speed $v(t)$ for the parameter values from Table~\ref{Tab:1}, except that now the parameter $d_1$ is doubled, $d_1=0.12$ [m]. It can be observed that the value of $d_1$ has a significant effect on the average speed $\bar{v}$, namely when decreasing $d_1$ from $d_1=0.12$ [m] (as shown in Fig.~\ref{fig2A}(d)) to $d_1=0.06$ [m] (as shown in Fig.~\ref{fig2A}(b)) the mean speed increases from about $\bar{v}\approx 2\cdot10^{-3}$ [m/s] to about $\bar{v}\approx 0.01$ [m/s]. We will further discuss the effect of $d_1$ on the mean speed of the steady-state symmetric periodic solution in Section~\ref{S:4}.

In Fig.~\ref{fig3A} we show a temporal simulation for the parameter values from Table~\ref{Tab:1}, but for a lower frequency of $\Omega=1.72$ [rad/s]. In Fig.~\ref{fig3A}(a) we show the dependence of the passive joint angle $\phi$ on time $t$. The solution is composed of the initial transient which converges to a periodic solution. Remarkably, unlike the previous case in Fig.~\ref{fig2A}, $\bar{\phi}$, which was defined in~\eqref{E:def means} does not vanish here, but its value is $\bar{\phi}\approx 1.0343$ [rad]. This means that the periodic solution is \emph{asymmetric} for this set of parameter values. In the inset of Fig.~\ref{fig3A}(a) we show that by choosing different initial conditions it is possible to obtain in this case a symmetric solution as well (shown by dashed purple curve). However,  a small deviation from these initial conditions yields a solution (shown by a solid green curve) which initially stays close to the symmetric solution, but then leaves the symmetric solution and after a sufficiently long time it converges to the asymmetric solution shown in panel (a). Thus, in this case the symmetric solution is \emph{unstable}, whereas the asymmetric solution is \emph{stable}.   

In Fig.~\ref{fig3A}(b) we show the evolution in time of the forward speed $v(t)$. It can be seen that after the initial transient the speed converges  to a periodic function with non-vanishing mean $\bar{v}$ (marked by a red dashed line) but, contrary to the previous case, its frequency is equal to the frequency of the actuation function $\psi(t)$. In Fig.~\ref{fig3A}(c) we show the dependence of the angle $\theta$ on time $t$. It can be observed that it converges to a function whose mean $\bar{\theta}(t)$ (marked by a red dashed line) is a linear increasing function of $t$. This means that the vehicle has nonzero net rotation during its forward motion.

In Fig.~\ref{fig3A}(d) we show the solution trajectory in the \linebreak $xy-$plane in two cases: the frequency of $\Omega=1.35$ [rad/s] (as in the rest of panels in Fig.~\ref{fig3A}) and the frequency of $\Omega=1.72$ [rad/s] (as in Fig.~\ref{fig2A}). The initial position in both cases is the same and it is marked with `x'. It is possible to see that while in the symmetric case after an initial transient the net motion of the Twistcar is along a straight line, in the asymmetric case after an initial transient the net motion of the Twistcar is along a circular arc.

\begin{figure}%[h]
	\centering
	\includegraphics[width=0.5\textwidth]{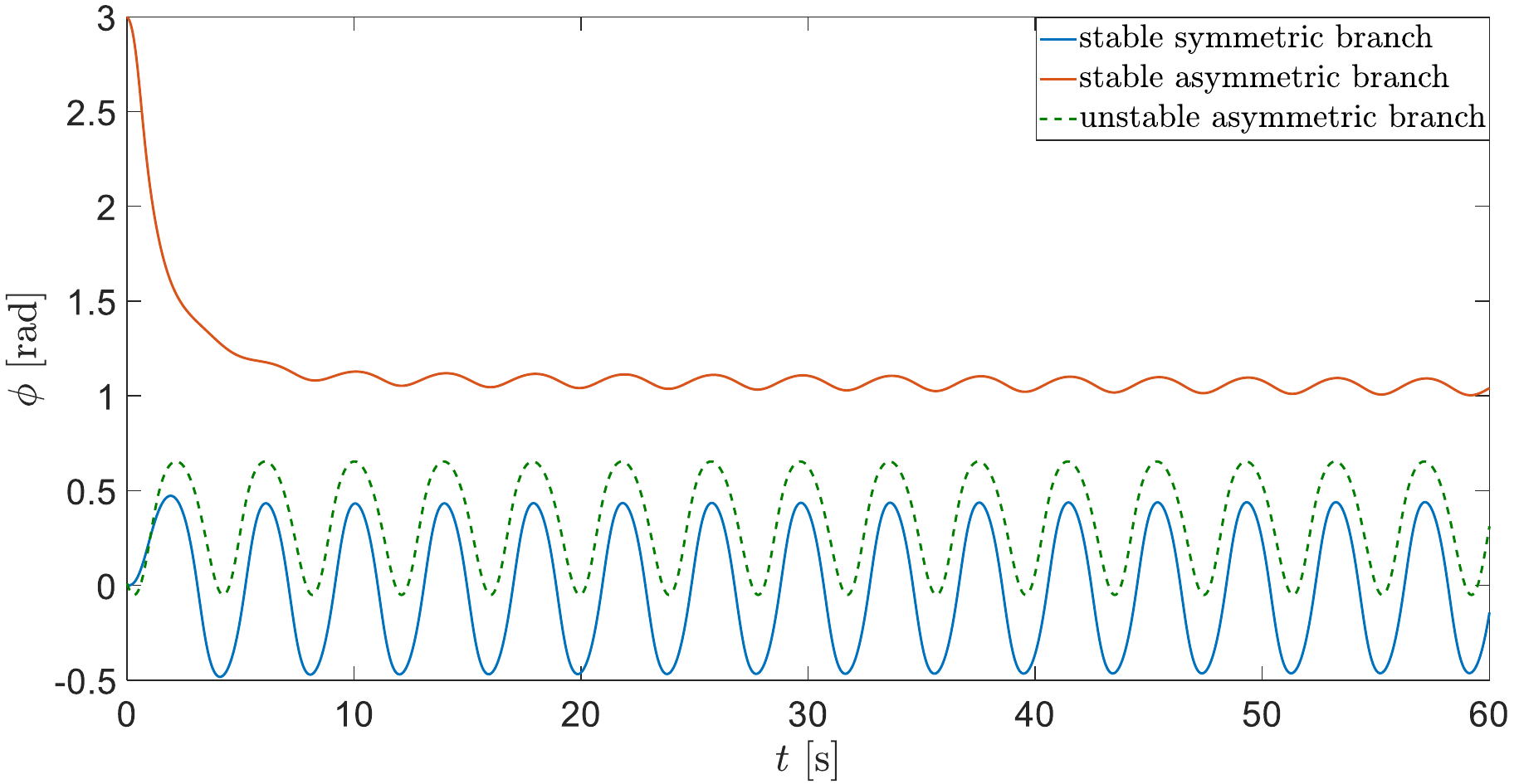}
	\caption{Numerical simulation for ``intermediate'' frequency $\Omega=1.6$ [rad/s] resulting in convergence to either a stable symmetric periodic solution (marked by a solid blue curve) or a stable asymmetric periodic solution (marked by a solid orange curve) depending on the initial conditions, where the physical parameters are given in Table~\ref{Tab:1}. There also exists an unstable asymmetric periodic solution (marked by a dashed green curve), which is obtained for appropriate initial conditions. In all three cases we show the passive angle $\phi$ versus time $t$.} \label{fig4A}
\end{figure}

\begin{figure}%[h]
	\centering
	\includegraphics[width=0.5\textwidth]{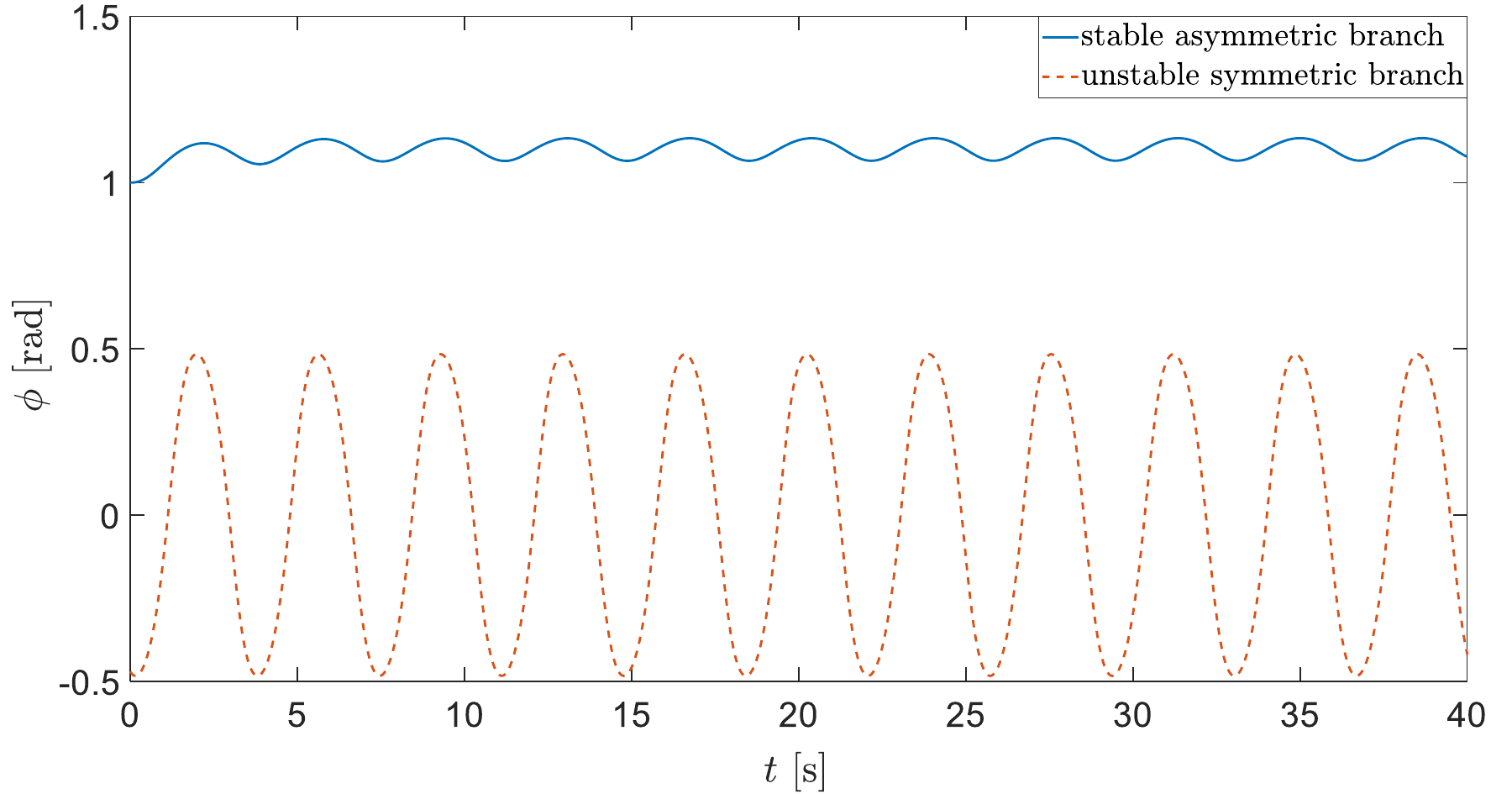}
	\caption{Numerical simulation for ``high'' frequency $\Omega=1.72$ [rad/s], which is the same case as for Fig.~\ref{fig2A} except that now $d_1=0.04$ [m], where the rest of physical parameters are given in Table~\ref{Tab:1}. The passive angle $\phi$ versus time $t$ in two cases: convergence to a stable asymmetric periodic solution with $\bar{\phi}\neq 0$ (shown by solid blue curve) and an unstable symmetric periodic solution with $\bar{\phi}=0$ (shown by dashed orange curve).} \label{fig5A}
\end{figure}

Since in the case of ``high'' frequency $\Omega=1.72$ [rad/s] (as shown in Fig.~\ref{fig2A}) the simulation converges to a symmetric solution and in the case of ``low'' frequency $\Omega=1.35$ [rad/s] (as shown in Fig.~\ref{fig3A}) the simulation converges to an asymmetric solution, in Fig.~\ref{fig4A} we show the dependence of the passive joint angle $\phi$ on time $t$ for an intermediate frequency of  $\Omega=1.6$ [rad/s], where the rest of parameter values are the same as previously (as given in Table~\ref{Tab:1}). It can be seen that in this case the simulation may converge to either a stable symmetric periodic solution or a stable asymmetric periodic solution depending on the initial conditions. Moreover, in this case there exists also an unstable asymmetric periodic solution, which is obtained when the initial conditions are chosen appropriately. Thus, we may conclude that the value of frequency $\Omega$ affects the existence of stable/unstable symmetric and asymmetric periodic solutions.

An interesting question which arises in view of the results shown in Figs.~\ref{fig2A}--\ref{fig4A} is if only the value of frequency $\Omega$ affects the existence of stable/unstable symmetric and asymmetric periodic solutions or any other parameter values, say the location of rotor's COM $d_1$, may also affect this issue. To answer this question, we show in Fig.~\ref{fig5A} the dependence of $\phi$ on time $t$ for the same frequency, of $\Omega=1.72$ [rad/s], and the same physical parameters as in Fig.~\ref{fig2A} (see Table~\ref{Tab:1}), except $d_1$ which is now $d_1=0.04$ [m]. While in Fig.~\ref{fig2A} we obtain convergence to a symmetric periodic solution, in Fig.~\ref{fig5A} we show convergence to an asymmetric periodic solution, where also an unstable symmetric periodic solution does exist and may be obtained with an appropriate choice of the initial conditions. Hence, our conclusion is that also the location of rotor's COM $d_1$ affects the existence of stable/unstable symmetric and asymmetric periodic solutions. 

We have found numerically that this nonlinear system has many interesting dynamical phenomena of multiplicity of periodic solutions, and changes of their properties depending on actuation frequency and other parameters. The aim of the following sections is to formulate these phenomena analytically, using asymptotic approximations.

%To investigate the dependence of the existence of stable/unstable symmetric and asymmetric periodic solutions on $\Omega$ and $d_1$, in Section~\ref{S:Reduction of the system} we will obtain a simpler system of 3 non-linear ODEs which governs the Twistcar's dynamics. Then in Section~\ref{S:Analysis of periodic solutions} we will perform numerical investigation regarding the existence of stable/unstable symmetric and asymmetric periodic solutions as a function of $\Omega$ and $d_1$, and find numerically the bifurcation points and stability transition. Finally, in Section~\ref{S:4} we will derive an asymptotic approximation for the Twistacr's motion 
%and in Section~\ref{S:44}, by using harmonic balance, we will obtain an approximation for the bifurcation between symmetric and asymmetric solutions.

\section{Reduction of the system's dynamics}\label{S:Reduction of the system}
Here we present reorganization of the equations of motion in a more reduced form.
%To reduce the system, we transfer the velocity to the body-frame system. 
First, we transform the generalized velocities to body-frame velocities \linebreak $\textbf{v}_b = \begin{bmatrix}v_{\parallel} & v_{\perp} & \dot{\theta} & \dot{\phi} & \dot{\psi}\end{bmatrix}^T$, using the transformation matrix $\textbf{R}_b$, as follows
%First, we assume that the vector of the body-frame velocity can be generalized as follows, 
\begin{equation} \label{general: body-frame transform}
	\begin{aligned}
		&\mathbf{v}_b = \mathbf{R}_b(\theta)^T \mathbf{\dot{q}} \text{, where }
		%&\mathbf{v} = (v_x, v_y, \dot{\theta}, \dot{\phi},\dot{\psi})^T\\
		\mathbf{R}_b(\theta) = 
		\begin{bmatrix}
			\cos{(\theta)} & -\sin{(\theta)} & 0 & 0 &0 \\
			\sin{(\theta)} & \cos{(\theta)} & 0 & 0 &0\\
			0 & 0 & 1 & 0 &0\\
			0 & 0 & 0 & 1 &0\\
			0& 0  &0 &0 &1
		\end{bmatrix}.
	\end{aligned}
\end{equation}
Expressing the constraints in~\eqref{world frame constraint equations} in terms of the body-frame velocities $\textbf{v}_b$, we have that
\begin{equation} \label{Constraints Reduced}
	\begin{aligned}
	\textbf{W} \dot{\textbf{q}} &= \underbrace{\left( \textbf{W} \textbf{R}_b \right)}_{\textbf{W}_b} \underbrace{\left( {\textbf{R}_b}^{-1} \dot{\textbf{q}} \right)}_{\textbf{v}_b}\\
	&= \begin{bmatrix} 0 && 1 && 0 && 0 && 0\\ -\sin{\phi} && \cos{\phi} && l_1 \cos{\phi} - l_2 && -l_2 && 0 \end{bmatrix} \begin{bmatrix} v_{\parallel} \\ v_{\perp} \\ \dot{\theta} \\ \dot{\phi} \\ \dot{\psi} \end{bmatrix} = \textbf{0}.
	\end{aligned}
\end{equation}

Next, we reduce the number of the body-frame velocities to the vector of 3 unknowns $\textbf{v}_r = \begin{bmatrix}v_{\parallel} & \dot{\theta} & \dot{\psi}\end{bmatrix}^T$. Note that the skid velocity of point $P_1$, $v_{\perp}$, is identically zero according to our definition and the non-holonomic constraints. Hence, the transformation from the reduced velocities $\textbf{v}_r$ to body-frame velocities $\textbf{v}_b$ may be expressed as,
\begin{equation} \label{E:reduction}
	\textbf{v}_b = \textbf{S}_r \textbf{v}_r, \quad \text{where} \quad \textbf{S}_r =
	\begin{bmatrix}
		1 & 0 & 0\\
		0 & 0  & 0\\
		0 & 1 & 0\\
		-\frac{\sin{\phi}}{l_2} & \frac{l_1}{l_2} \cos{\phi} - 1 & 0 \\
		0 & 0 & 1
	\end{bmatrix}.
\end{equation}

Note that body-frame velocities $\textbf{v}_b$ in~\eqref{E:reduction} automatically satisfy the nonholonomic constraints in~\eqref{Constraints Reduced}. From the transformation in~\eqref{general: body-frame transform} and the reduction in~\eqref{E:reduction}, we obtain a relation between the generalized velocities and the reduced body velocities, $\dot{\textbf{q}} = \textbf{S} \textbf{v}_r$, where the transformation matrix is given by $\textbf{S} = \textbf{R}_b \textbf{S}_r$. Substituting this relation between generalized and reduced velocities into \eqref{organized constraints Lagrange Equation}, we obtain the reduced equation 
\begin{equation} \label{Reduced Dynamic Equation}
	\textbf{M}_r(\phi) \dot{\textbf{v}}_r + \textbf{B}_r(\phi, \textbf{v}_r) + \textbf{D}_r(\phi, \textbf{v}_r) = \textbf{F}_r,
\end{equation}
where
\begin{equation*}
	\begin{aligned}
	&\textbf{M}_r(\phi) = \textbf{S}^T \textbf{M} \textbf{S} \in \mathbb{R}^{3 \times 3}, \\
	&\textbf{B}_r(\phi, \textbf{v}_r) = \textbf{S}^T \textbf{M} \dot{\textbf{S}} \textbf{v}_b + \textbf{S}^T \textbf{B} \in \mathbb{R}^{3 \times 1}, \quad \textbf{F}_r = \begin{bmatrix}0 & 0 & \tau\end{bmatrix}^T,
	\end{aligned}
\end{equation*}
and where $\textbf{D}_r=\textbf{S}^T\textbf{D}(\textbf{q},\dot{\textbf{q}})$, after substituting the velocity relations~\eqref{general: body-frame transform} and~\eqref{E:reduction}. Detailed expressions for $\textbf{M}_r$, $\textbf{B}_r$, and $\textbf{D}_r$ appear in Appendix A, equations~\eqref{E:M_r}-\eqref{E:D_r}.

Next, we express the equations in~\eqref{Reduced Dynamic Equation} and the constraints in~\eqref{Constraints Reduced} in a more concise way. We start by using the first constraint in~\eqref{Constraints Reduced}, according to which $v_{\perp}$ vanishes. Hence, for brevity, we denote $v=v_{\parallel}$. Note that in this case the system in~\eqref{Reduced Dynamic Equation} contains three equations, where the last equation is decoupled from the rest. Since $\psi(t)$ is a given function of time, the dynamics of the system is governed by the first two equation in~\eqref{Reduced Dynamic Equation}, whereas the third equation contains the actuation torque $\tau$.
%, so that we may use $\ddot{\theta}$ and $\ddot{\psi}$ in order to express the input torque $\tau$. 
Thus, the system in~\eqref{Reduced Dynamic Equation} reduces to the following two equations in three unknowns $(v, \dot{\theta}, \phi)$,
\begin{subequations}\label{E:dimensional problem}
	\begin{align}
		&m_r\dot{v}-d_1 m_r \dot{\theta}^2+c\left[(2+\cos^2{(\phi)})v+\frac{l_1}{2}\sin{(2\phi)}\dot{\theta}\right]=0,\\
		&(I_r+d_1^2 m_r)\ddot{\theta}+I_r \ddot{\psi}+d_1 m_r v\dot{\theta} \notag\\
		&\phantom{(I_r+d_1^2 m_r)\ddot{\theta}}+c\left[\frac{l_1}{2}v+(2d^2+l_1^2\sin^2{(\phi)})\dot{\theta}\right]=0.
	\end{align}
%\end{subequations}
The third equation is obtained from the second constraint in~\eqref{Constraints Reduced}, which yields that
\begin{equation} \label{4th Dynamic Equation}
	\dot{\phi} = -\frac{1}{l_2} v \sin{\phi} + \dot{\theta} \left( \frac{l_1}{l_2} \cos{\phi} - 1 \right).
\end{equation}
\end{subequations}
%For further details see equations~\eqref{E:M_r}--\eqref{E:D_r} in Appendix.

Next, to render the problem in~\eqref{E:dimensional problem} dimensionless, we first define $t_c = m_r / c$ as the characteristic time scale of the problem with $c$ denoting the rolling dissipation coefficient. Then, all time-dependent quantities are scaled as
\begin{equation} \label{eq16}
	\begin{aligned}
	&\tilde{v} = \frac{t_c}{l_1}v , \quad \dot{\tilde{v}} = \frac{d \tilde{v}}{d \tilde{t}} = \frac{{t_c}^2}{l_1}\dot{v} , \quad \dot{\tilde{\theta}} =  t_c\dot{\theta}, \quad \ddot{\tilde{\theta}} = {t_c}^2\ddot{\theta} , \\
	 &\dot{\tilde{\phi}} =  t_c\dot{\phi}, \quad \ddot{\tilde{\psi}} = {t_c}^2\ddot{\psi}, \quad \tilde{t} = \frac{t}{t_c}, \quad \omega = t_c\Omega.  
	 \end{aligned}
\end{equation}
For convenience, we define the following dimensionless parameters,
\begin{equation}\label{E:dimless numbers}
	\alpha=\frac{s}{l_1}, \quad \beta=\frac{l_2}{l_1}, \quad \delta=\frac{d_1}{l_1}, \quad \eta=\frac{I_r}{m_rl_1^2},
\end{equation}
and hereafter remove the tilde ($\,\,\tilde{}\,\,$) symbol
from all of the variables, where we use the convention that they represent scaled dimensionless quantities. 

Finally, substituting~\eqref{eq16} and~\eqref{E:dimless numbers} into the system in~\eqref{E:dimensional problem} and denoting for brevity $\sigma=\dot{\theta}$, we obtain the following dimensionless system of nonlinear ODEs,
\begin{equation}\label{E:basic system}
	\dot{\mathbf{z}}=\left(
	\begin{aligned}
		&\frac{1}{\beta}\left[-v\sin{(\phi)}+\left(-\beta+\cos{(\phi)}\right)\sigma\right]\\
		&-\frac{\left[2\eta\ddot{\psi}+v\sin{(2\phi)}+\left(\alpha_1-\cos{(2\phi)}+2\delta v\right)\sigma\right]}{2(\delta^2+\eta)}\\
		&\delta\sigma^2-0.5\sigma\sin{(2\phi)}-0.5(5+\cos{(2\phi)})v,
	\end{aligned}
	\right)
\end{equation}
where $\alpha_1=1+4\alpha^2$, $\mathbf{z}=(\phi,\sigma,v)^T$ and $\ddot{\psi}=-A\omega^2\sin{(\omega t)}$.

\section{Analysis of periodic solutions}\label{S:Analysis of periodic solutions}
In this section, we investigate periodic solutions and their stability by using the Poincar\'{e} map~\cite{Guckenheimer}. The methodology for finding and analyzing the periodic solutions is as follows. Note that in~\eqref{E:basic system}, the vector $\mathbf{z}=(\phi,\sigma,v)^T$ describes the system's solution. Next, we define $\mathbf{z}_k=\mathbf{z}(t=kt_p)$, $k=0,1,2,\ldots$, where $t_p=2\pi/\omega$ is the nondimensional period time. Since the actuation is periodic in $t_p$, it is possible to express the discrete dynamic equation, without time dependency, as
\begin{equation}
	\mathbf{z}_{k+1}=\mathbf{P}(\mathbf{z}_k),
\end{equation}
where $\mathbf{P}$ denotes the Poincar\'{e} map of the system, see~\cite{Guckenheimer} for further details. More specifically, we refer here to Poincar\'{e} stroboscopic map due to the use of the actuation's period as the sampling time interval. It is possible to evaluate $\mathbf{P}(\mathbf{z}_k)$ by numeric integration of~\eqref{E:basic system} over a period. Since a periodic solution of~\eqref{E:basic system} satisfies $\mathbf{z}(t)=	\mathbf{z}(t+t_p)$, in order to find it one needs to solve the system $\mathbf{P}(\mathbf{z}^*)=\mathbf{z}^*$. We find solution of $\mathbf{z}^*$ numerically, by using `fsolve' function in Matlab. 

The stability of a periodic solution is determined via the eigenvalues $\lambda_i$ of the linearization matrix
\begin{equation}\label{E:numerical stab crit}
	\mathbf{J}=\frac{d\mathbf{P}}{d\mathbf{z}}\Bigl|_{(\mathbf{z}=\mathbf{z}^*)}.
\end{equation}

Specifically, a periodic trajectory is asymptotically stable if and only if all of the eigenvalues, called Floquet multipliers~\cite{Guckenheimer}, satisfy $|\lambda_i|<1$. It is possible to estimate the matrix $\mathbf{J}$ by numerical differentiation of $\mathbf{P}$. Thus, by using Poincar\'{e} map it is possible to find periodic solutions and assess their stability.

\begin{figure*}[ht!]
	\centering
	\includegraphics[width=\textwidth]{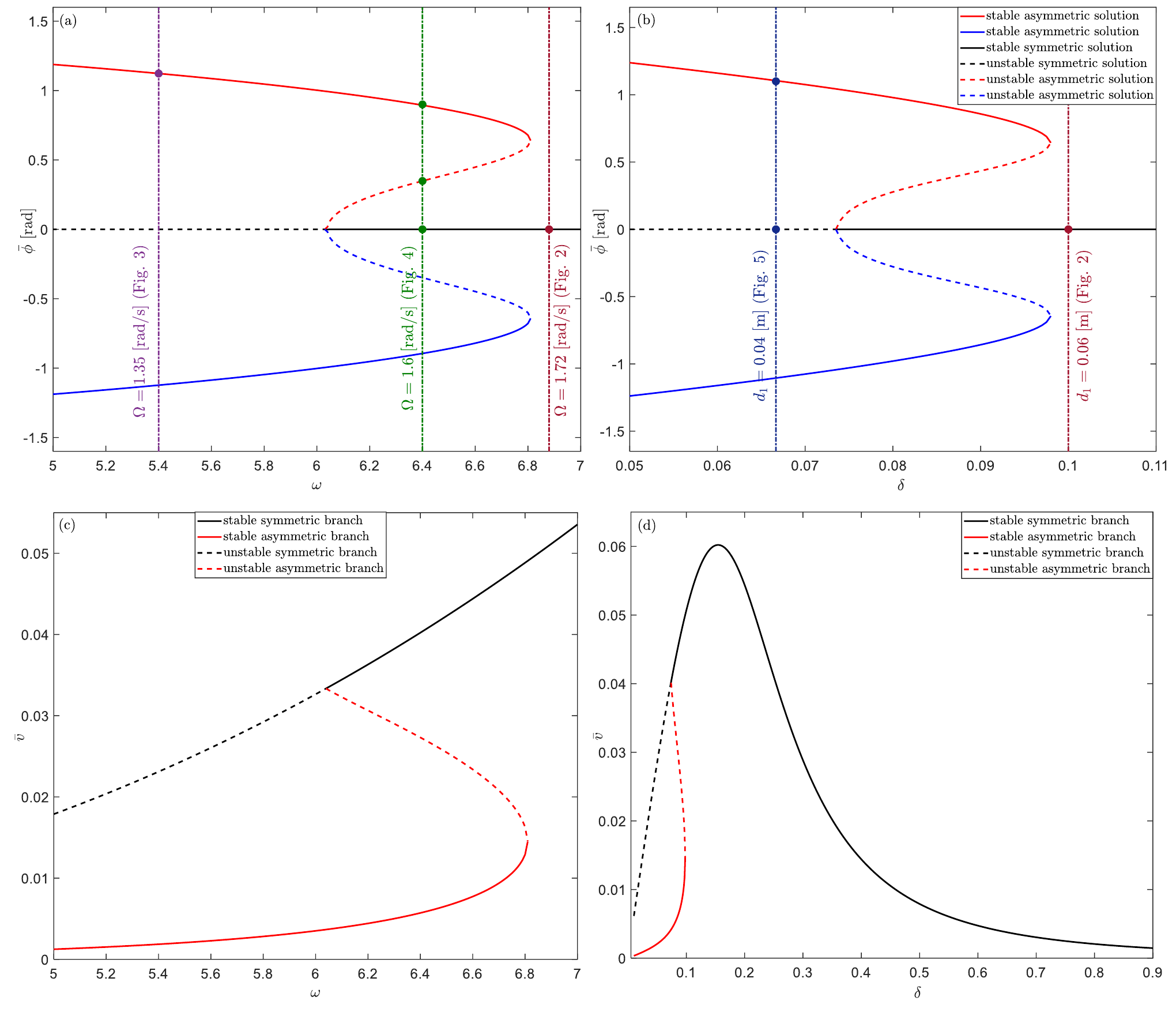}
	\caption{Effects of nondimensional frequency and rotor's COM location on periodic solutions' properties, additional asymmetric solution branches, and bifurcations. Parameter values are given in Table~\ref{Tab:1} and nondimensional scaling in~\eqref{E:dimless numbers}. (a) Mean angle $\bar{\phi}$ versus frequency $\omega$. (b) Mean angle $\bar{\phi}$ versus rotor's COM location $\delta$ for frequency of $\omega=6.88$. The three dashed-dotted vertical lines in panel (a) correspond to the results shown in Figs.~\ref{fig2A}--\ref{fig4A} and the two dashed-dotted vertical lines in panel (b) correspond to Figs.~\ref{fig2A} and~\ref{fig5A}. (c) Mean speed $\bar{v}$ versus frequency $\omega$. (d) Mean speed $\bar{v}$ versus rotor's COM location $\delta$ for  frequency of $\omega=6.88$. The legend in panel (b) refers to both panels (a)-(b).} \label{fig2}
\end{figure*}

In Fig.~\ref{fig2} we show multiplicity of the periodic solutions and their stability obtained by using Poincar\'{e} map. More specifically, in Fig.~\ref{fig2}(a) we show the mean angle $\bar{\phi}$ versus frequency $\omega$ for parameter values given in Table~\ref{Tab:1}, where solid curves indicate stable solutions and dashed curves indicate unstable solutions. It can be seen that for sufficiently small $\omega$, $\omega<6.03$, there exist two stable asymmetric solutions and an unstable symmetric solution. Then, at $\omega=6.03$ there exists a \emph{pitchfork bifurcation}, where the symmetric solution becomes stable and two additional asymmetric branches, which are unstable, emerge. At $\omega=6.81$ there exists an additional bifurcation, known as \emph{fold bifurcation}, which means that the two stable asymmetric solutions exist for $\omega<6.81$ and the two unstable asymmetric solutions exist for $6.03<\omega<6.81$. Note that all values are truncated to two significant digits.

In Fig.~\ref{fig2}(b) we show the mean angle $\bar{\phi}$ versus the scaled rotor's COM position $\delta=d_1/l_1$, for $\omega=6.88$ and the rest of parameter values as given in Table~\ref{Tab:1}. It can be seen when the parameter $\delta$ is varied, there also exists a pitchfork bifurcation at $\delta \approx 0.0735$, where the symmetric periodic solution loses its stability and two additional unstable asymmetric periodic solutions emerge. Moreover, there also exists a fold-bifurcation at $\delta\approx0.098$, which distinguishes between the ranges of existence and non-existence of asymmetric solutions.

In Figs.~\ref{fig2}(c) and (d) we show the mean speed $\bar{v}$ versus $\omega$ and $\delta$ for that same parameter values as in In Figs.~\ref{fig2}(a) and (b), respectively. It is possible to observe that the mean speed of the symmetric periodic solution is always greater or equal (at the bifurcation point) than the corresponding mean speed of the asymmetric periodic solution. Moreover, while the mean speed of the symmetric periodic solution is monotonically increasing with the frequency $\omega$, it has an optimal value as a function of $\delta$. Therefore, there exists an optimal position for the rotor's COM, which is approximately 20\% of the main body length from the rear link, which yields the maximal mean forward speed.

\begin{figure}%[h]
	\centering
	\includegraphics[width=0.5\textwidth]{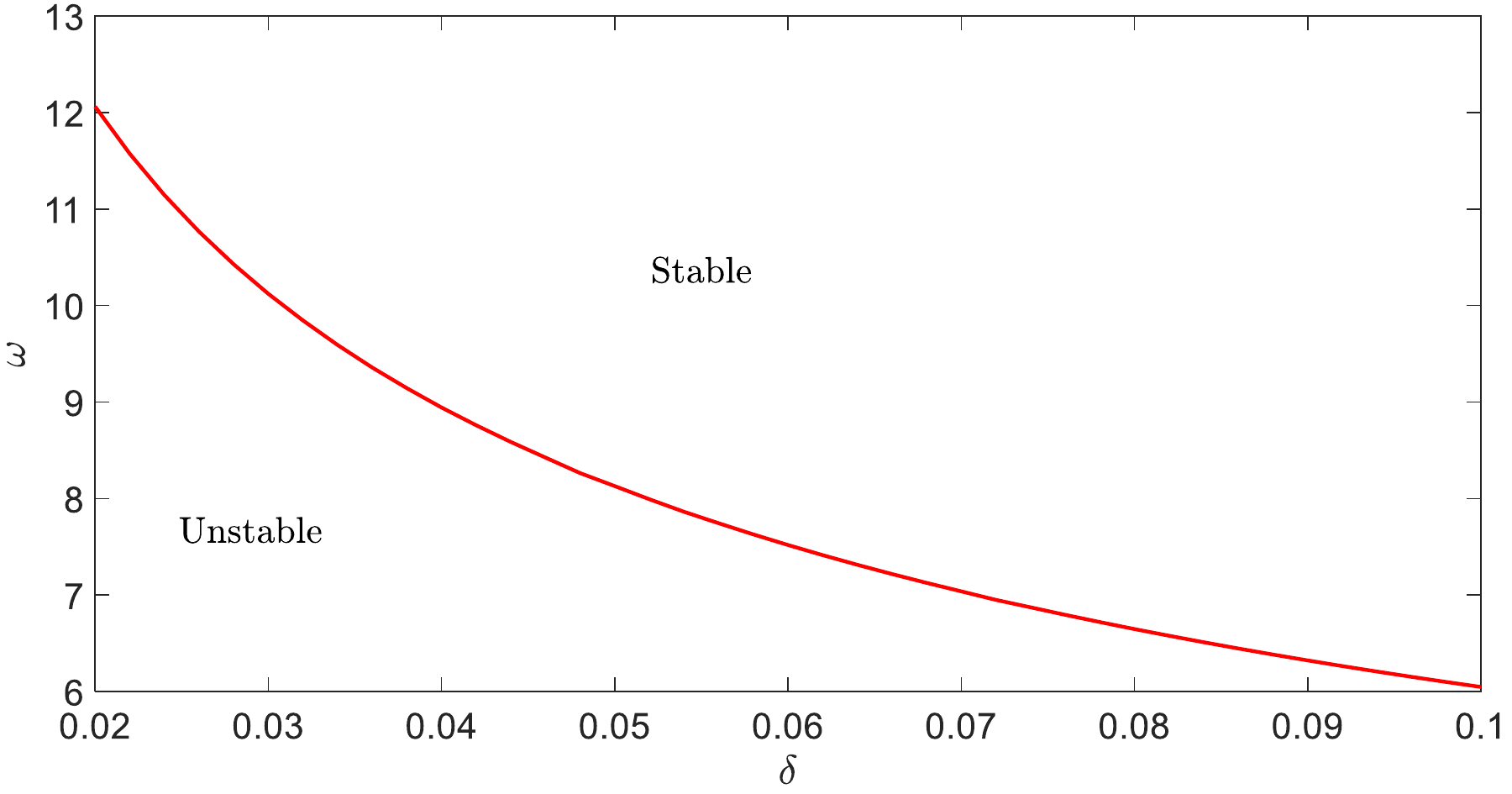}
	\caption{Stability transition curve for the symmetric periodic solution. The bifurcation frequency $\omega$ versus rotor's COM location $\delta$, for the parameter values given in Table~\ref{Tab:1} and nondimensional scaling in~\eqref{E:dimless numbers}.} \label{fig6A}
\end{figure}

In Fig.~\ref{fig6A} we show the stability transition diagram for the symmetric periodic solutions in the $(\delta\omega)-$plane. The curve in the figure was obtained numerically by using Poincar\'{e} map and the bisection method. More specifically, for any fixed $\delta$, by modifying the values of $\omega$, we follow the symmetric periodic solution and calculate its Floquet multipliers. To obtain the stability transition point we use the bisection method, which allows us to find $\omega$ corresponding to the point where the maximal eigenvalue magnitude crosses 1. We repeat this procedure for a discrete sequence of values of $\delta$ in the desired interval. 

As shown in Fig.~\ref{fig6A}, the symmetric periodic solution is stable above the stability transition curve, and unstable below it. Moreover, it can be seen that the stability transition curve may be expressed via a function of bifurcation frequency $\omega$ versus $\delta$, which is monotonically decreasing. One of our aims in the coming sections is to derive a closed-form polynomial which will approximate this stability transition curve.   

%\section{Asymptotic analysis}\label{sec.ana}

\section{Asymptotic analysis of symmetric periodic solution}\label{S:4}
We now present asymptotic analysis of symmetric periodic solution of the problem in~\eqref{E:basic system}. Under the assumption of sufficiently small $A$, $\eta$, and $\omega$, let us define a small parameter $0<\varepsilon=A\eta\omega^2\ll1$. Thus, we assume the following expansions
\begin{equation}\label{E:asymptotic expansions}
	\begin{aligned}
		&\phi(t)=\phi_0(t)+\varepsilon \phi_1(t)+O(\varepsilon^2),\\
		&\sigma(t)=\sigma_0(t)+\varepsilon \sigma_1(t)+O(\varepsilon^2),\\
		&v(t)=v_0(t)+\varepsilon v_1(t)+\varepsilon^2 v_2(t)+O(\varepsilon^3).
	\end{aligned}
\end{equation} 
Note that $\phi_0(t)=\sigma_0(t)=v_0(t)=0$, because when there is no actuation the solution is identically zero. 
%(This may be proven e.g., by expanding the system in~\eqref{E:basic system} about $\phi=0$ up to order $O(\varepsilon)$, then using that $\bar{\phi}_0=0$ by definition of symmetry and thus $\dot{\phi}_0=1/\beta(-\beta+1)\sigma_0$ yielding that $\bar{\sigma}_0=\bar{v}_0=0$, and then harmonic balance at any order implies that $\sigma_0(t)=v_0(t)=0$ and hence $\phi_0(t)=0$).

To find the first-order correction, we start from the second equation in~\eqref{E:basic system}, which leads at order $O(\varepsilon)$, to
\begin{equation}\label{E:eq first order sigma}
	\dot{\sigma}_1=\frac{\sin{(\omega t)}}{\delta^2+\eta}-q\sigma_1, \quad \text{where} \quad q=\frac{2\alpha^2}{\delta^2+\eta}.
\end{equation}
%where we denote for brevity
%Equation~\eqref{E:eq first order sigma} may be expressed as 
%\begin{equation}
%	\frac{\partial}{\partial t}\left(e^{qt}\sigma_1\right)=e^{qt}  \frac{\sin{(\omega t)}}{\delta^2+\eta},
%\end{equation}
%whose integration with respect to $t$, yields the following general solution
The general solution of~\eqref{E:eq first order sigma} is given by
\begin{equation}\label{E:gen sol sig}
	\begin{aligned}
	\sigma_1(t)&=e^{-qt}\left[\frac{\omega}{(\delta^2+\eta)(q^2+\omega^2)}+C\right]\\
	&+\frac{\omega \cos{(\omega t)}-q\sin{(\omega t)}}{(\delta^2+\eta)(q^2+\omega^2)}. 
	\end{aligned}
\end{equation}
%where $C$ is an arbitrary constant of integration. Now we use periodicity requirement which implies that
%\begin{equation}
%	C=-\frac{\omega}{(\delta^2+\eta)(q^2+\omega^2)},
%\end{equation}
The first term in~\eqref{E:gen sol sig} decays exponentially in time and thus in steady-state the periodic solution may be expressed as,
%leading that the periodic and steady-state solution, which is obtained after the exponential decay, is given by
\begin{equation}\label{E:sol sigma_1 symm}
	\sigma_1(t)=b_1\sin{(\omega t)}+b_2\cos{(\omega t)},
\end{equation}
where we define for brevity
\begin{equation}\label{E:b1 and b2}
	b_1=\frac{q}{(\delta^2+\eta)(q^2+\omega^2)}, \quad b_2=-\frac{\omega}{(\delta^2+\eta)(q^2+\omega^2)}. 
\end{equation}

Substituting the solution in~\eqref{E:sol sigma_1 symm} into the first equation in~\eqref{E:basic system}, we obtain:
\begin{equation}
	\dot{\phi}_1(t)=\frac{1-\beta}{\beta}\sigma_1(t),
\end{equation}
which yields that the steady-state symmetric solution is given by,
\begin{equation}\label{E:sol phi sym}
		\phi_1(t)=a_1\sin{(\omega t)}+a_2\cos{(\omega t)},
\end{equation}
where we define for brevity
\begin{equation}\label{E:a1 and a2}
	a_1=\frac{(\beta-1)}{\beta(q^2+\omega^2)(\delta^2+\eta)}, \quad 	a_2=\frac{q(\beta-1)}{\beta\omega(q^2+\omega^2)(\delta^2+\eta)}.
\end{equation}

The expressions for $\phi_1(t)$ and $\sigma_1(t)$ are then substituted into the third equation in~\eqref{E:basic system}, and we obtain at order 
$O(\varepsilon)$ that $\dot{v}_1(t)=-3v_1(t)$. Hence, $v_1(t)=0$ in steady-state after the transient response has decayed. 
Thus, expanding up to order $O(\varepsilon^2)$, we obtain that
\begin{equation}\label{E:v1 and v2}
	v_1(t)=0, \quad \dot{v}_2=\delta\sigma_1^2-\sigma_1\phi_1-3v_2. 
\end{equation}

Solving this equation we obtain that, in steady-state after the transient response has decayed, the solution is given by
\begin{equation}\label{E:sol v sym}
	v_2(t)=c_0+c_1\sin{(2\omega t)}+c_2\cos{(2\omega t)},
\end{equation}
where we define for brevity
\begin{equation}
	 c_1=\frac{3q_1+2q_2\omega}{9+4\omega^2}, \quad c_2=\frac{3q_2-2q_1\omega}{9+4\omega^2},
\end{equation}
where
\[
\begin{aligned}
	&c_0=(\delta b_1^2  -a_1 b_1  +\delta b_2^2  -a_2 b_2)/6,\\
	&q_1=\delta b_1 b_2 -0.5(a_2 b_1+a_1 b_2),\\
	&q_2= 0.5(\delta b_2^2  -a_2 b_2  +a_1 b_1 -\delta b_1^2).
\end{aligned}
\]

It is easy to verify by substituting $a_i$ from~\eqref{E:a1 and a2} and $b_i$ from~\eqref{E:b1 and b2}, that that $c_0$ may be expressed as
\begin{equation}
	c_0=\frac{\delta}{6\bigl[4\alpha^4+(\delta^2+\eta)^2\omega^2\bigr]}.
\end{equation}
Note that from~\eqref{E:sol v sym} it follows that $\bar{v}_2=c_0$, and hence the leading-order approximation for the mean speed is given by
\begin{equation}\label{E:bar v}
	\begin{aligned}
		\bar{v}&=\varepsilon^2 c_0 +O(\varepsilon^3) \\
		&=\frac{A^2\eta^2\omega^4\delta}{6\bigl[4\alpha^4+(\delta^2+\eta)^2\omega^2\bigr]}+O(\varepsilon^3).
	\end{aligned}
\end{equation}
When considering this expression as a function of $\delta$, it is possible to conclude that $\bar{v}$ attains at most one maximum (in the physically meaningful range $0<\delta<1$), which is located at
\begin{equation}\label{E:delta opt}
	\delta_{\text{opt}}=\sqrt{\frac{\sqrt{12\alpha^4+4\eta^2\omega^2}}{3\omega}-\frac{\eta}{3}}.
\end{equation}
Thus, the maximal mean speed is
\begin{equation}\label{E:v_mean_opt}
	\bar{v}_{\text{max}}=\frac{A^2\eta^2\omega^4\delta_{\text{opt}}}{6\bigl[4\alpha^4+(\delta_{\text{opt}}^2+\eta)^2\omega^2\bigr]}.
\end{equation}

\begin{figure*}[ht!]
	\centering
	\includegraphics[width=\textwidth]{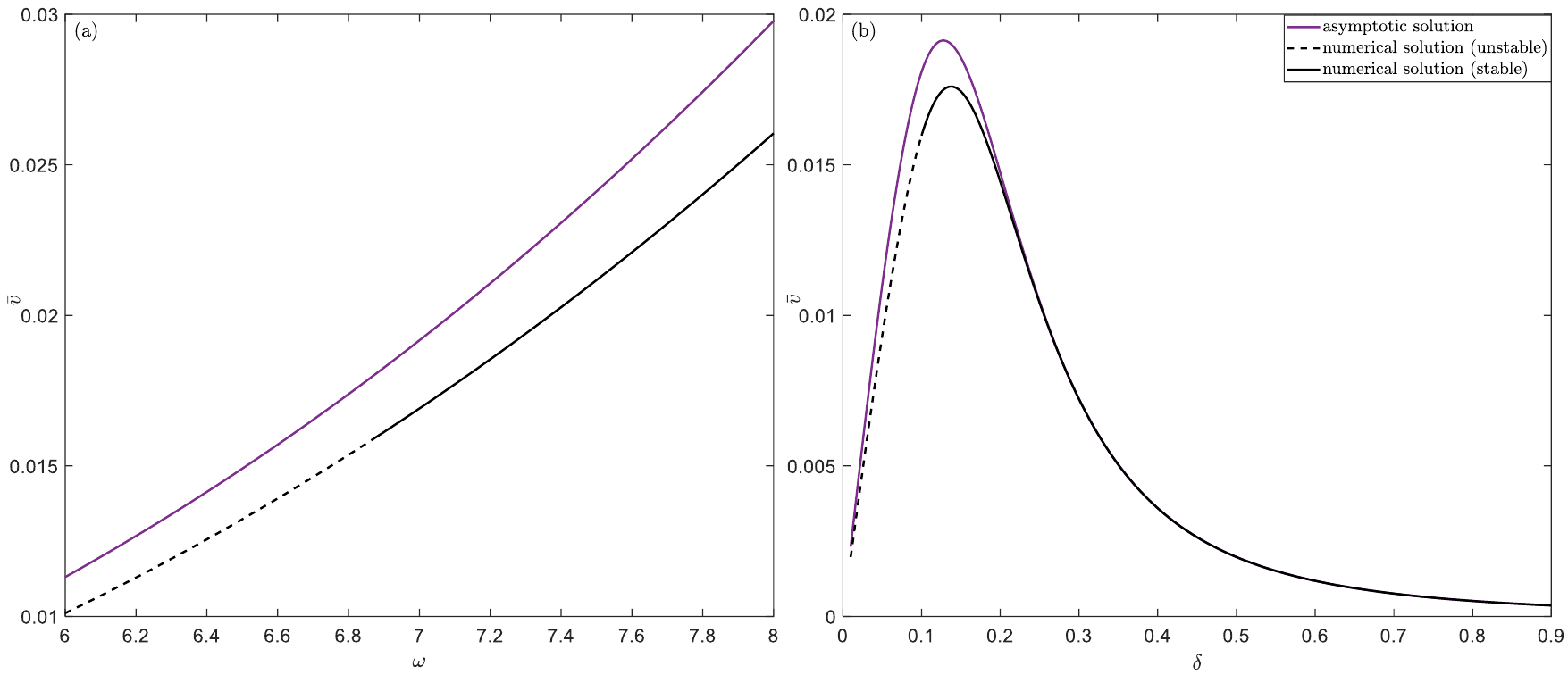}
	\caption{Comparison between asymptotic (purple curve) and numerical (black curve) results for (a) mean speed $\bar{v}$ versus frequency $\omega$ and (b) mean speed $\bar{v}$ versus rotor's COM location $\delta$. The dashed and solid black curves correspond to the unstable and stable symmetric periodic solutions, respectively. The parameter values are as given in Table~\ref{Tab:1} and nondimensional scaling in~\eqref{E:dimless numbers}, except the amplitude $A$, which is now $A=0.5$, and where in (b) $\omega=6.88$.} \label{fig8}
\end{figure*}

\begin{figure*}[ht!]
	\centering
	\includegraphics[width=\textwidth]{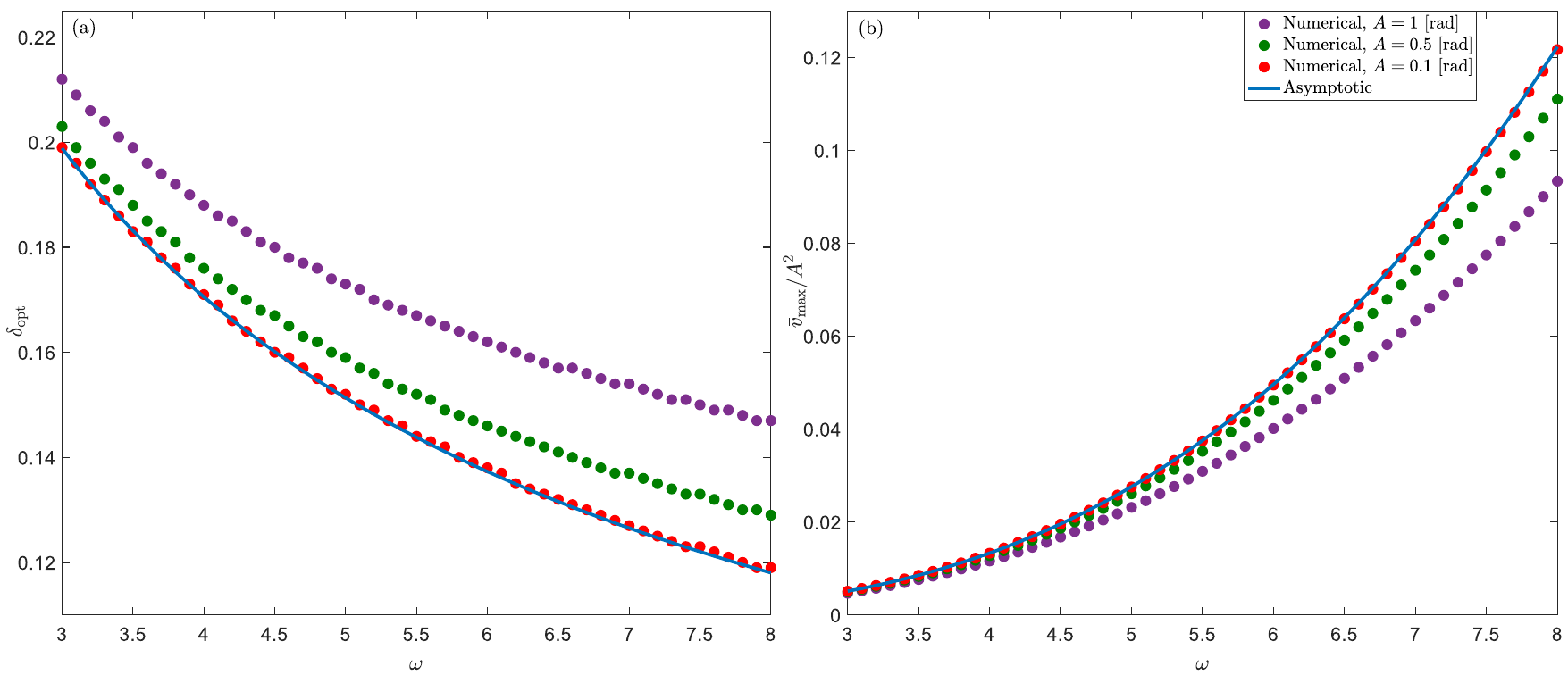}
	\caption{Comparison between asymptotic (solid blue curve) and numerical (red, green, and purple circles) results for (a) optimal rotor's COM location $\delta_{\text{opt}}$ versus frequency $\omega$ and (b) optimal mean speed scaled by $A^2$, namely $\bar{v}_{\text{max}}/A^2$ versus frequency $\omega$. The physical parameter values are as given in Table~\ref{Tab:1} and nondimensional scaling in~\eqref{E:dimless numbers}, except the amplitude $A$, which is now $A\in\{0.1,\,0.5,\,1\}$. The asymptotic results in (a) and (b) were obtained according to~\eqref{E:delta opt} and~\eqref{E:v_mean_opt}, respectively.} \label{fig9}
\end{figure*}

In Fig.~\ref{fig8}(a,b) we show a comparison between numerical and asymptotic results calculated by using the approximate formula in~\eqref{E:bar v} for the mean speed $\bar{v}$ versus frequency $\omega$ for $\delta=0.1$, and versus $\delta$ for $\omega=6.88$, respectively. The rest of physical parameters are given in Table~\ref{Tab:1}, except for the actuation amplitude which is $A=0.5$. It can be seen that in both cases there is a very good agreement between numerical and asymptotic results, which will become even better upon decreasing the value of $A$. 
%The mean speed $\bar{v}$ is a monotone increasing function of $\omega$, but it is not monotone function of $\delta$, and as it is expected according to equation~\eqref{E:delta opt} it attains maximum at $\delta_{\text{opt}}\approx0.128$.

In Fig.~\ref{fig9}(a) we show a comparison between numerical and asymptotic results calculated by using the formula in~\eqref{E:delta opt} for $\delta_{\text{opt}}$ versus $\omega$, for the parameter values given in Table~\ref{Tab:1} and three values of actuation amplitude $A$. Note that while the asymptotic expression for $\delta_{\text{opt}}$ is independent of $A$, the numerically calculated $\delta_{\text{opt}}$ depends on $A$. As $A$ (and thus $\varepsilon$) decreases, the numerical results get closer to the asymptotic ones, so that for $A=0.1$ they nearly coincide. According to the results shown in the graph, $\delta_{\text{opt}}$ is a monotonically decreasing function of $\omega$, which means that in order to obtain optimal speed for higher frequency, one needs to move the rotor's COM closer to the rear wheels (see Fig.~\ref{fig1}).

In Fig.~\ref{fig9}(b) we show a comparison between numerical and asymptotic results calculated by using the formula in~\eqref{E:v_mean_opt} for $\bar{v}_{\text{max}}/A^2$ versus $\omega$, for the parameter values given in Table~\ref{Tab:1} and three values of $A$. It is possible to see that the agreement between the numerical and asymptotic results gets better as $A$ and $\omega$ decrease, since $\varepsilon$ is proportional to $A\omega^2$. In particular, for $A=0.1$ the numerical and asymptotic results nearly coincide.  
%Moreover, in all three cases, the optimal mean speed is a monotone increasing function of frequency. Thus, the results in Fig.~\ref{fig9} indicate that in order maximize the optimal speed (for a given amplitude) one should maximize the frequency and move the rotor's COM close to the rear wheels, where the approximate optimal distance, scaled by the main link length, between the rotor's COM and the rear wheels is given by the equation in~\eqref{E:delta opt}.    

The comparison between numerical and asymptotic results demonstrates that the asymptotic analysis performed in this section provides a good approximation for the symmetric periodic solution. However, asymptotic analysis alone cannot capture the stability transition of the symmetric periodic solution, and thus in the next section we shall develop a methodology that allows to achieve this.

\section{Bifurcations analysis of periodic solutions}\label{S:44}
In this section we derive an analytic approximation for a stability transition condition of the symmetric periodic solution, where the methodology is based on harmonic balance~\cite{Nayfeh_Mook_2008}. 

Note that if we consider the asymptotic expansions  of the steady-state solution for $0<\varepsilon\ll1$, which are given in~\eqref{E:asymptotic expansions} in Section~\ref{S:4}, it is possible to obtain the epsilon-order of all terms in the equations in~\eqref{E:basic system}. More specifically, the leading-order terms in the first and second equations in~\eqref{E:basic system} are of order $O(\varepsilon)$, whereas the leading-order terms in the third equation in~\eqref{E:basic system} are of order $O(\varepsilon^2)$. Thus,  considering the leading-order equation resulting from the second equation in~\eqref{E:basic system}, we conclude that the term $2\delta v\sigma$ may be neglected, since it is of higher order $O(\varepsilon^3)$, assuming $\delta=O(1)$. Neglecting the term $2\delta v\sigma$ in the second equation in~\eqref{E:basic system} and multiplying the result by $2\left(\delta^2+\eta\right)$, gives that
\begin{equation}\label{E:second neglect}
\begin{aligned}
	&2\left(\delta^2+\eta\right)\dot{\sigma}\\
	&\phantom{2\left(\delta^2\right)}=-\left[\left(2\eta\ddot{\psi}+v\sin{\left(2\phi\right)}\right)+\left(1+4\alpha^2-\cos{\left(2\phi\right)}\right)\sigma\right].
\end{aligned}
\end{equation}

Next, we multiply the first equation in~\eqref{E:basic system} by 2$\beta\cos{\left(\phi\right)}$, which yields that
\begin{equation}\label{E:first eq multiplied}
	2\beta\dot{\phi}\cos{\left(\phi\right)}=-v\sin{\left(2\phi\right)}{+2\left(-\beta\cos{\left(\phi\right)}+\cos^2{\left(\phi\right)}\right)\sigma}.
\end{equation}
Then, we subtract equation~\eqref{E:second neglect} from equation~\eqref{E:first eq multiplied}, which results in the following equation,
\begin{equation}\label{E:linear sig}
	\begin{aligned}
	2\beta\dot{\phi}\cos{\left(\phi\right)}&-2\left(\delta^2+\eta\right)\dot{\sigma}\\
	&=2\eta\ddot{\psi}+\left(-2\beta\cos{\left(\phi\right)}+4\alpha^2+2\right)\sigma.
	\end{aligned}
\end{equation}
Importantly, equation~\eqref{E:linear sig} has only two unknowns, $\phi(t)$ and $\sigma(t)$, and moreover it is linear in $\sigma(t)$. Now, replacing the second equation in the system~\eqref{E:basic system} by~\eqref{E:linear sig}, we obtain the following system,
\begin{equation}\label{E:basic system simplified}
	\begin{aligned}
		&\dot{\phi}=\frac{1}{\beta}\left[-v\sin{(\phi)}+\left(-\beta+\cos{(\phi)}\right)\sigma\right],\\
		&2\beta\dot{\phi}\cos{\left(\phi\right)}-2\left(\delta^2+\eta\right)\dot{\sigma}=2\eta\ddot{\psi}\\
		&\phantom{2\beta\dot{\phi}\cos{\left(\phi\right)}}+\left(-2\beta\cos{\left(\phi\right)}+4\alpha^2+2\right)\sigma,\\
		&\dot{v}=\delta\sigma^2-0.5\sigma\sin{(2\phi)}-0.5(5+\cos{(2\phi)})v.
	\end{aligned}
\end{equation}
Next, we expand $\sin{(\phi)}$ and $\cos{(\phi)}$ in~\eqref{E:basic system simplified} as power series in $\phi$, truncated to low orders, and obtain that
\begin{subequations}\label{E:basic system simplified expanded}
	\begin{align}
		&\dot{\phi}=\frac{1}{\beta}\left[-v\left(\phi-\frac{\phi^3}{6}\right){+\left(-\beta+1-\frac{\phi^2}{2}\right)\sigma}\right]+O(\phi^4), \label{E:basic system simplified expanded1}\\
		&	2\beta\dot{\phi}\left(1-\frac{\phi^2}{2}+\frac{\phi^4}{24}\right)-2\left(\delta^2+\eta\right)\dot{\sigma} \notag\\
		&\phantom{2\beta\dot{\phi}\left(1\right)}=2\eta\ddot{\psi}+\left(-2\beta+\beta\phi^2-\frac{\beta\phi^4}{12}+4\alpha^2+2\right)\sigma \notag\\
		&\phantom{2\beta\dot{\phi}\left(1\right)}+O(\phi^6), \label{E:basic system simplified expanded2}\\
		&\dot{v}=\delta\sigma^2-0.5\sigma\left(2\phi-\frac{4}{3}\phi^3\right)-3v+O(\phi^4). \label{E:basic system simplified expanded3}
	\end{align}
\end{subequations} 
Note that unlike the expansion in orders of $\varepsilon$ conducted above in Section~\ref{S:4}, here we only expand the trigonometric expressions in $\phi$. The orders of expansion for each expression in~\eqref{E:basic system simplified expanded} were chosen as the minimal ones that enable obtaining the bifurcation condition.

Next, we use a method of harmonic balance~\cite{Nayfeh_Mook_2008}, which allows us to obtain analytical expression for the pitchfork bifurcation.  This method is based on the assumption that the periodic solution for the steady-state of the system in~\eqref{E:basic system simplified expanded} is a series of harmonic functions. Moreover, according to our numerical results (see Fig.~\ref{fig2A}(a)-(c)), $\phi(t)$ and $\sigma(t)$ are approximately single harmonic functions of $\omega$, whereas $v$ has the first two harmonic functions in $\omega$. Thus, neglecting higher order harmonics, we may assume the following harmonic series solution:
\begin{equation}\label{E:harmonic balance}
	\begin{aligned}
		&\phi(t)=\tilde{a}_0+\tilde{a}_1\sin{\left(\omega t\right)}+\tilde{a}_2\cos{\left(\omega t\right)},\\
		&\sigma(t)=\tilde{b}_0+\tilde{b}_1\sin{\left(\omega t\right)}+\tilde{b}_2\cos{\left(\omega t\right)},\\
		&v(t)=\tilde{c}_0+\tilde{c}_1\sin{\left(\omega t\right)}+\tilde{c}_2\cos{\left(\omega t\right)}+\tilde{c}_3\sin{\left(2\omega t\right)}\\
		&\phantom{v(t)}+\tilde{c}_4\cos{\left(2\omega t\right)}.		
	\end{aligned}
\end{equation} 
By substituting the expansions given in equation~\eqref{E:harmonic balance} into \eqref{E:basic system simplified expanded} and equating the harmonics prefactors, one obtains a system of 11 algebraic equations with 11 unknowns, $\tilde{a}_i$, $\tilde{b}_i$, with $i=0,1,2$, and $\tilde{c}_i$ with $i=0,1,2,3,4$. 
%Now we demonstrate that the neglect of the term $2\delta v\sigma$ allows us to eliminate the coefficients $\tilde{b}_i$, with $i=0,1,2$, and $\tilde{c}_i$ with $i=0,1,2,3,4$, resulting in a much simpler algebraic system with only three equations for coefficients $\tilde{a}_i$, $i=0,1,2$, as the only unknowns.

Substituting~\eqref{E:harmonic balance} into equation~\eqref{E:basic system simplified expanded2} and equating the coefficients of the first order harmonics, we get that the vector $\mathbf{b}=(\tilde{b}_0,\tilde{b}_1,\tilde{b}_2)^T$ may be expressed as a function of the vector $\mathbfit{a}=(\tilde{a}_0,\tilde{a}_1,\tilde{a}_2)^T$ as follows,
\begin{equation}\label{E:M_b def}
	\mathbf{M}_b(\mathbfit{a}) \mathbf{b} =\mathbf{F}_b(\mathbfit{a}),
\end{equation}
where $\mathbf{M}_b(\mathbfit{a})$ and $\mathbf{F}_b(\mathbfit{a})$ depend on the physical parameters of the problem and $\mathbfit{a}$, see equations~\eqref{E:M_b} and~\eqref{E:F_b} in Appendix B. %\red{(To get these expression explicitly, run the code ``symbolic-bessel-151225.m'' in Matlab and see ``Mat-b'' and ``F-b''.)} 
Thus, we may conclude that there exists a rational function $\mathbf{B}$, so that 
\begin{equation}\label{E:b-a}
	\mathbf{b}=\mathbf{B}(\mathbfit{a}),
\end{equation}
where $\mathbf{B}(\mathbfit{a})=\mathbf{M}_b^{-1}(\mathbfit{a})\mathbf{F}_b(\mathbfit{a})$.

Now, since the equation in~\eqref{E:basic system simplified expanded3} is linear in $v(t)$, substituting~\eqref{E:harmonic balance} into it, equating the coefficients of the first and second order harmonics, and using~\eqref{E:b-a}, we get that
\begin{equation}\label{E:M_c def}
	\mathbf{M}_c \mathbf{c} =\mathbf{F}_c(\mathbfit{a},\mathbf{b}),
\end{equation}
%\begin{equation}
%	\begin{pmatrix} 
%		3	& 0 & 0 & 0 & 0\\
%		0   & (\mathbf{M}^{(1,2)}_c)_{1,1} & (\mathbf{M}^{(1,2)}_c)_{1,2} & 0 &0\\
%		0	& (\mathbf{M}^{(1,2)}_c)_{2,1} & (\mathbf{M}^{(1,2)}_c)_{2,2} & 0 &0\\
%		0	&0	&0	& (\mathbf{M}^{(3,4)}_c)_{1,1} & (\mathbf{M}^{(3,4)}_c)_{1,2}\\
%		0	&0	&0	& (\mathbf{M}^{(3,4)}_c)_{2,1} & (\mathbf{M}^{(3,4)}_c)_{2,2}\\
%	\end{pmatrix}	
%	\begin{pmatrix}c_0\\
%		c_1\\
%		c_2\\
%		c_3\\
%		c_4\end{pmatrix}=\begin{pmatrix}
%		(C_4)_2\\
%		(\mathbf{F}^{(1,2)}_c)_1\\
%		(\mathbf{F}^{(1,2)}_c)_2\\
%		(\mathbf{F}^{(3,4)}_c)_1\\
%		(\mathbf{F}^{(3,4)}_c)_2
%		\end{pmatrix},
%\end{equation}
where $\mathbf{c}=(\tilde{c}_0,\tilde{c}_1,\tilde{c}_2,\tilde{c}_3,\tilde{c}_4)^T$. Note that $\mathbf{M}_c$ depends only on $\omega$, whereas $\mathbf{F}_c$ depends on $\mathbfit{a}$, $\mathbf{b}$, $\omega$ and the physical parameters of the problem. For more details, see equations~\eqref{E:M_c1} and~\eqref{E:F_c} in Appendix B.
% \red{where the matrices $\mathbf{M}^{(1,2)}_c$, $\mathbf{M}^{(3,4)}_c$, the arrays $\mathbf{F}^{(1,2)}_c$, $\mathbf{F}^{(3,4)}_c$, and the constant $(C_4)_2$ can be obtained explicitly by running the code ``symbolic-bessel-151225.m'' in Matlab and looking for ``Mat-c12'', ``Mat-c34'', ``F-c12'', ``F-c34'' and ``C4''.}
Thus, we may conclude that there exists a rational function $\mathbf{C}$, so that 
\begin{equation}\label{E:c-a}
	\mathbf{c}=\mathbf{C}(\mathbfit{a}),
\end{equation}
where $\mathbf{C}(\mathbfit{a})=\mathbf{M}_c^{-1}\mathbf{F}_c(\mathbfit{a},\mathbf{b}=\mathbf{B}(\mathbfit{a}))$. Note that the linear dependence of $\mathbf{b}$ on $\mathbfit{a}$ in~\eqref{E:M_b def} and of $\mathbf{c}$ on $\mathbfit{a}$ in~\eqref{E:M_c def} is a simplification enabled only due to negligence of the term $2\delta v\sigma$ from the second equation in system~\eqref{E:basic system}.

In the last step, we substitute~\eqref{E:harmonic balance} into equation~\eqref{E:basic system simplified expanded1}, equate the coefficients of the first order harmonics, and use \eqref{E:b-a} and~\eqref{E:c-a}. Thus, we obtain a non-linear algebraic system of 3 equations,
\begin{equation}\label{E:harmonic balance nonlinear sys}
	\mathbf{F}(\mathbfit{a})=0.
\end{equation} 
where $\mathbf{F}$ is $3\times1$ vector, denoted by $(F_1(\mathbfit{a}), F_2(\mathbfit{a}), F_3(\mathbfit{a}))^T$. 
$F_1$ is obtained from comparing coefficients for the free (DC) term in the harmonic balance equation~\eqref{E:basic system simplified expanded1}. It is proven in equations~\eqref{E:gen form a0}--\eqref{E:form c_i} in Appendix B that the equation $F_1(\mathbfit{a})=0$ takes the form,
\begin{equation}\label{E:a0}
	\tilde{a}_0 \Upsilon(\tilde{a}^2_0,\tilde{a}_1,\tilde{a}_2)=0.
\end{equation}
%For further details, see equations~\eqref{E:gen form a0}--\eqref{E:form c_i} in Appendix B. 
Equation~\eqref{E:a0} typically has two possible types of solutions for the unknowns $\tilde{a}_0$, $\tilde{a}_1$, and $\tilde{a}_2$. The first type of solutions is with $\tilde{a}_0=0$, which corresponds to symmetric periodic solution of $\phi(t)$ according to~\eqref{E:harmonic balance}. The second type of solutions is with $\Upsilon(\tilde{a}^2_0,\tilde{a}_1,\tilde{a}_2)=0$ and $\tilde{a}_0^2 \neq 0$, which corresponds to a pair of asymmetric periodic solutions. The pitchfork bifurcation point occurs when the two types of solutions coincide, which means
\begin{equation}\label{E:bifurc cond}
	\tilde{a}_0=0 \quad \text{and} \quad \Upsilon(0,\tilde{a}_1,\tilde{a}_2)=0.
\end{equation} 
This can be interpreted as a bifurcation condition on $\omega$ and system parameters.

The bifurcation condition in~\eqref{E:bifurc cond}, determines the parameter values where the symmetric periodic solution loses its stability. Hence, the bifurcation condition in~\eqref{E:bifurc cond} can be used in order to determine stability transition curve. 
In what follows, we present two methods that can be used to find the stability transition curve of the symmetric periodic solutions in the $(\delta\omega)-$plane, in addition to the numerical calculation shown in Fig.~\ref{fig6A}, where the remaining parameters are fixed and given in Table~\ref{Tab:1}.

In the first way, which we refer as to ``numerical HB solution," we find numerically the coefficients $\tilde{a}_1$ and $\tilde{a}_2$, which satisfy the bifurcation condition in~\eqref{E:bifurc cond}. More specifically, for any fixed $\delta$, we use `fsolve' function in Matlab and solve the following system of equations in the unknowns $\tilde{a}_1$, $\tilde{a}_2$, and $\omega$:
\begin{equation}
	\left\{\begin{aligned}
		&\Upsilon(0,\tilde{a}_1,\tilde{a}_2,\omega)=0,\\
		&F_2(0,\tilde{a}_1,\tilde{a}_2,\omega)=0,\\
		&F_3(0,\tilde{a}_1,\tilde{a}_2,\omega)=0.
	\end{aligned}\right.
\end{equation}
In particular, the solution of this system determines $\omega$ as a function of $\delta$ (and other parameters of the problem), for which the stability transition occurs.

\begin{figure*}[ht!]
	\centering
	\includegraphics[width=\textwidth]{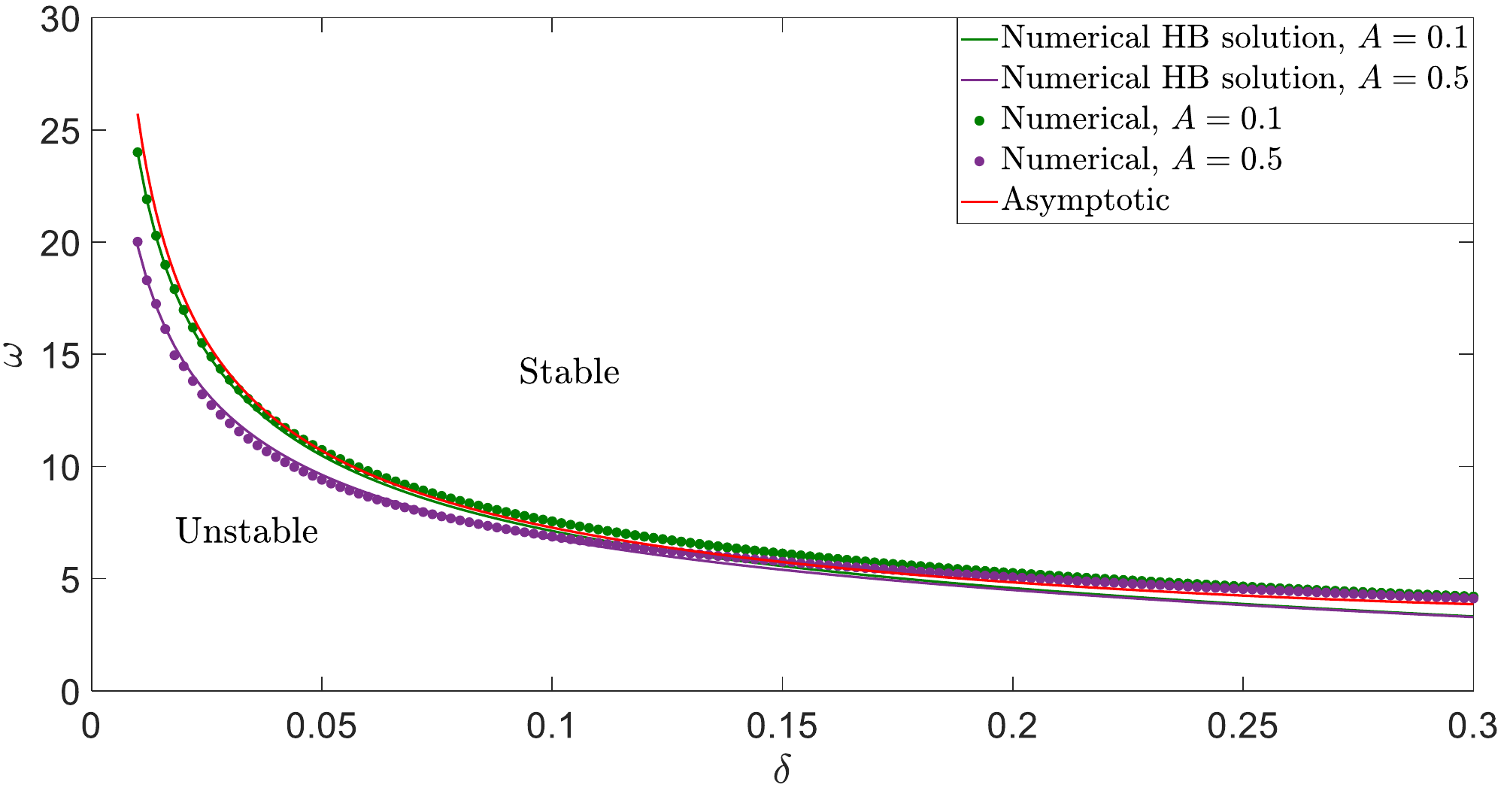}
	\caption{Stability transition curve obtained on $(\delta\omega)-$plane. Numerical HB solution for the amplitude of $A\in\{0.1,0.5\}$ [rad] which satisfies the criterion defined in~\eqref{E:bifurc cond}, is denoted by green and purple solid curves, respectively. Numerical calculation of the stability transition curve by finding symmetric periodic solution (i.e finding fixed point of the Poincar\'{e} map) and calculating its Floquet multipliers (as in equation~\eqref{E:numerical stab crit} and graph in Fig.~\ref{fig6A}) for the amplitude of $A\in\{0.1,0.5\}$ [rad] is denoted by green and purple circles. Asymptotic solution denoted by red solid curve, was obtained by finding the roots of the closed-form polynomial given in~\eqref{E:closed form pol}, and is approximately independent of $A$ in the range $A\in[0.1,0.5]$. Above the stability transition curve, the symmetric periodic solution is stable, and below it is unstable.} \label{fig10}
\end{figure*}

In the second way, which we entitle ``asymptotic," we find an analytic approximation for the stability transition curve of the symmetric periodic solution, by using the perturbation expansion solution from equation~\eqref{E:sol phi sym}. More specifically, we substitute $a_1$ and $a_2$, given in~\eqref{E:a1 and a2}, into the equation $\Upsilon(0,\tilde{a}_1,\tilde{a}_2,\omega)=0$ namely, we substitute $\tilde{a}_i = \varepsilon a_i$ for $i=1,2$, where $\varepsilon=A\eta\omega^2$ and $a_i$ are given in~\eqref{E:a1 and a2}.
This gives a high-degree polynomial in $\omega$, where the rest of parameters are fixed. This equation is an approximation of the bifurcation condition in~\eqref{E:bifurc cond}. In order to reduce the complexity of this equation, we further approximate $a_1$ and $a_2$, by defining another small parameter $\tilde{\varepsilon}=\sqrt{{(\delta}^2+\eta)}\ll1$ and assuming that $\eta\ll \delta$ so that $\eta=O\left(\tilde{\varepsilon}^2\right)$, 
and $\alpha=O\left(\sqrt{\tilde{\varepsilon}}\right)$. Hence, if we define 
\begin{equation}\label{E:scaling alpha}
	\tilde{\alpha}=\frac{\alpha}{\sqrt{\tilde{\varepsilon}}}, \quad \tilde{\eta}=\frac{\eta}{\tilde{\varepsilon}^2}, 
\end{equation}
we may expand $a_1$ and $a_2$ given in~\eqref{E:a1 and a2}, in $\tilde{\varepsilon}$ as follows
\begin{equation}\label{E:a1 and a2 expanded}
	\begin{aligned}
		&a_1=\frac{\left(-1+\beta\right)}{4{\tilde{\alpha}}^4\beta}-\frac{\left(\omega^2\left(-1+\beta\right)\right)\tilde{\varepsilon}^2}{16\left({\tilde{\alpha}}^8\beta\right)}+O\left(\tilde{\varepsilon}^4\right),\\ &a_2=\frac{\beta-1}{2\omega{\tilde{\alpha}}^2\beta\tilde{\varepsilon}}-\frac{\omega\tilde{\varepsilon}\left(\beta-1\right)}{8{\tilde{\alpha}}^6\beta}+\frac{\omega^3\left(-1+\beta\right)\tilde{\varepsilon}^3}{32{\tilde{\alpha}}^{10}\beta}+O\left(\tilde{\varepsilon}^4\right).
	\end{aligned}
\end{equation}
Substituting the assumptions in~\eqref{E:scaling alpha} and~\eqref{E:a1 and a2 expanded} into the numerator of (the rational) equation $\Upsilon(0,\tilde{a}_1,\tilde{a}_2,\omega)=0$, and expanding the numerator of the resulting function in $\tilde{\varepsilon}$ and $\varepsilon$ up to orders $O(\tilde{\varepsilon}^3)$ and $O(\varepsilon^3)$, respectively, we obtain that the stability transition curve may be approximated by a polynomial in two variables $\delta$ and $\omega$,
\begin{equation}\label{E:closed form pol}
	\begin{aligned}
		d_8(\delta)\omega^8+d_6(\delta)\omega^6+d_4(\delta)\omega^4+d_2(\delta)\omega^2+d_0(\delta)=0,
	\end{aligned}
\end{equation}
where the rest of parameters are given in Table~\ref{Tab:1}, and $d_8$, $d_6$, $d_4$, $d_2$, and $d_0$ are polynomials in $\delta$. The complete calculation can be found in symbolic Matlab code, which is provided in the supplementary information (SI) \cite{SI}. In Appendix B (see~\eqref{E:closed form pol coeffs}), we give the numerical values of the coefficients for the physical parameters which are as given in Table~\ref{Tab:1}, nondimensional scaling in~\eqref{E:dimless numbers}, and amplitude $A=0.1$.
To obtain the bifurcation point $\omega(\delta)$, where the symmetric periodic solution loses its stability, we need to find the corresponding root of the bi-quartic polynomial in~\eqref{E:closed form pol}. For each $\delta$, there are up to four roots for $\omega^2$, where negative and complex-valued roots are discarded. The rest of roots (except one root, which we denote by $\omega(\delta)$) are far from the numerical solution of the harmonic balance equations. Thus, the red curve in Fig.~\ref{fig10} was obtained by plotting the roots $\omega(\delta)$, which are the closest to the numerical solution of the harmonic balance equations.
  
%\red{Note that as a part of Supplementary Files we provide a symbolic function in Matlab, entitled ``stability-transitions-explicit-polynomial,'' which allows to reproduce the process for obtaining a closed form polynomial that approximates the stability transition curves as a function of any parameters in our model.}   

In Fig.~\ref{fig10} we show the stability transition curve obtained by using the abovementioned three ways: the numerical one as shown in Fig.~\ref{fig6A}, Numerical HB, and Asymptotic. It can be seen that there is an excellent agreement between all of them, where for sufficiently small values of $\delta$ the numerical and Numerical HB solutions nearly coincide. The dependence of the stability transition curves (see, numerical and Numerical HB solutions) on the amplitude $A$ is relatively small, and it decreases as $\delta$ increases. Even though the asymptotic solution misses the dependence of the stability curve on the amplitude $A$ (both values of $A$, $A=0.1$ and $A=0.5$ [rad] yield the same curve), it still captures well the monotonic decrease of $\omega$ as a function of $\delta$ along the stability curve. The asymptotic curve is very close to both other methods, where for sufficiently large $\delta$ it approximates the numerical solution even better than Numerical HB method. The main advantage of the asymptotic solution is its simplicity, which allows to obtain a very good approximation for the stability curve via finding roots of a closed-form polynomial.

\section{Concluding remarks}
In this study we investigated the dynamics of a rotor-actuated Twistcar robot with passive steering joint. We formulated the system's nonlinear dynamic equations of motion and performed numerical simulations. Our results revealed that there exists an optimal
rotor's COM position which maximizes the mean speed of the robot. Moreover, upon varying the actuation frequency $\omega$ or rotor's COM position, we showed that there exist two bifurcation points of periodic solutions, one of the type ``pitchfork'' and the second of the type ``fold.'' For the frequency smaller than the ``pitchfork'' critical frequency, there exists a pair of stable asymmetric periodic solutions and an unstable primary symmetric periodic solution. When $\omega$ is increased beyond the ``pitchfork'' bifurcation point, the primary symmetric periodic solution becomes stable and a pair of unstable asymmetric periodic solutions emerge. Increasing the frequency even more, we achieve a ``fold" bifurcation, where the stable and unstable asymmetric periodic solutions meet. For frequency larger than the ``fold'' bifurcation value, there exists only a stable symmetric periodic solution. For a constant frequency, but varying rotor's COM location, we obtain a similar picture of periodic solutions' multiplicity and bifurcations as described above. Thus, by using Poincar\'{e} map we constructed a stability transition map of the primary symmetric periodic solution in the frequency and rotor's COM position plane.

In addition, we used asymptotic tools which allowed us to obtain approximate explicit expressions describing the behavior of the system.
In particular, the perturbation analysis yielded an approximate solution for the symmetric branch, which allowed to find an analytical expression for the optimal rotor's COM location. By using harmonic balance, we obtained an approximation for the bifurcation points
of pitchfork type. Finally, based on the combination between perturbation analysis and  harmonic balance we formulated an analytic approximation for the stability transitions criterion as a function of frequency and rotor's COM location. We compared between the analytical and numerical results and obtained a very good
agreement.

The results illustrate the rich dynamics observed in \linebreak periodically-actuated systems with a passive shape coordinate. In addition, our work proves the significant contribution of asymptotic analysis for finding explicit formulation of some of the system's dynamic phenomena. Future work will attempt to demonstrate some of the theoretical results in experiments using wheeled articulated robotic vehicles.

%Furthermore, we constructed a stability transition map of the primary symmetric solution in the frequency and amplitude
%plane.

\section*{Supplementary information}
Supplementary material related to this article can be found online at \url{https://yizhar.net.technion.ac.il/files/2025/06/SI-MATLAB-file-Anna-Z.zip}. The Matlab code for reproducing the results in Fig.~\ref{fig10} and in particular the calculation of the coefficients which appear in equation~\eqref{E:closed form pol}
is available electronically in the supplementary information (SI).

\section*{Acknowledgments}
We wish to warmly thank Prof. Oleg Gendelman for many fruitful discussions and insights.
The work of Y. O. has been supported by Israel Science Foundation under grant \linebreak no. 1382/23.\\

{\bf{Data availability}}
The datasets generated during and/or analysed during the current study are available from the corresponding author on reasonable request. 

\section*{Declarations}
{\bf{Conflict of interests}}
The authors declare that they have no conflicts of interest.

\onecolumn{
% \titleformat{\section}{\normalfont\bfseries}{}{0em}{}
%\appendix
%\renewcommand{\thesection}{Appendix A}
\section*{Appendix A}
%\section*{Appendix A}\label{App A}
\setcounter{equation}{0}
\renewcommand{\theequation}{A.\arabic{equation}}
Under the assumption of  $m_1=m_2=I_1=I_2=0$, the matrices in \eqref{Reduced Dynamic Equation} can be expressed explicitly as,
\begin{equation}\label{E:M_r}
	\textbf{M}_r = \begin{bmatrix}
		m_r & 0 & 0 \\
		0 & J_r + {b_1}^2 m_r & J_r \\
		0 & J_r & J_r
	\end{bmatrix},
\end{equation}
\begin{equation}\label{E:B_r}
	\textbf{B}_r = \begin{bmatrix}
		-b_1 m_r {\dot{\theta}}^2 \\
		b_1 m_r v \dot{\theta} \\
		0
	\end{bmatrix},
\end{equation}
and
\begin{equation}\label{E:D_r}
	\textbf{D}_r = c \begin{bmatrix}
		\left(2 + \cos^2{\phi}\right)v + \frac{l_1}{2} \sin{(2\phi) \dot{\theta}} \\
		\frac{l_1}{2} v + \left(2 s^2 + {l_1}^2 \sin^2{\phi}\right) \dot{\theta} \\ 
		0
	\end{bmatrix}.
\end{equation}

\section*{Appendix B}
\setcounter{equation}{0}
\renewcommand{\theequation}{B.\arabic{equation}}
Here we give all details of the derivations in Section~\ref{S:44}. Let us start from the matrix $\mathbf{M}_b(\mathbfit{a})$ appearing in equation~\eqref{E:M_b def}. Substituting the expansions given in~\eqref{E:harmonic balance} into equation~\eqref{E:basic system simplified expanded2}, it is possible to verify that is given by,
	\begin{equation}\label{E:M_b}
		\mathbf{M}_b=\begin{bmatrix}
			& m_{1,1} & m_{1,2}  & m_{1,3}\\
			& -m_{1,2} & m_{2,2}  & m_{2,3}\\
			& -m_{1,3} & m_{3,2}  & m_{3,3}\\
		\end{bmatrix},
		\end{equation}
		where
		\begin{equation}\label{E:M_b terms}
			\begin{aligned}
			&m_{1,1}=\beta\left(\tilde{a}_0^2 + \frac{\tilde{a}_1^2}{2} + \frac{\tilde{a}_2^2}{2}\right) - 2\beta -
			\frac{\beta}{96}(8\tilde{a}_0^4 + 24\tilde{a}_0^2 \tilde{a}_1^2 + 24 \tilde{a}_0^2 \tilde{a}_2^2 + 3\tilde{a}_1^4 + 6 \tilde{a}_1^2 \tilde{a}_2^2 + 3 \tilde{a}_2^4) + 4\alpha^2 + 2,\\
			&m_{1,2}=\tilde{a}_0 \tilde{a}_1 \beta - \frac{\tilde{a}_0 \tilde{a}_1 \beta}{24}(4\tilde{a}_0^2 + 3\tilde{a}_1^2 + 3\tilde{a}_2^2),\\
			&m_{1,3}=\tilde{a}_0 \tilde{a}_2\beta - \frac{\tilde{a}_0 \tilde{a}_2 \beta}{24}(4\tilde{a}_0^2 + 3\tilde{a}_1^2 + 3\tilde{a}_2^2),\\
			%&m_{b(2,1)}=\frac{\tilde{a}_0 \tilde{a}_1 \beta}{24}(4\tilde{a}_0^2 + 3\tilde{a}_1^2 + 3\tilde{a}_2^2) - \tilde{a}_0 \tilde{a}_1\beta,\\
			&m_{2,2}=-2\alpha^2 + \beta + \frac{\beta}{192}(8\tilde{a}_0^4 + 36\tilde{a}_0^2 \tilde{a}_1^2 + 12 \tilde{a}_0^2 \tilde{a}_2^2 + 5\tilde{a}_1^4 + 6\tilde{a}_1^2 \tilde{a}_2^2 + \tilde{a}_2^4) - \beta\left(\frac{\tilde{a}_0^2}{2} + \frac{3\tilde{a}_1^2}{8} + \frac{\tilde{a}_2^2}{8}\right) - 1,\\
			&m_{2,3}=\omega(\delta^2 + \eta) - \frac{\tilde{a}_1 \tilde{a}_2 \beta}{4} + \frac{\tilde{a}_1 \tilde{a}_2 \beta}{48}(6\tilde{a}_0^2 + \tilde{a}_1^2 + \tilde{a}_2^2),\\
			&m_{3,2}=\frac{\tilde{a}_1 \tilde{a}_2 \beta}{48}(6\tilde{a}_0^2 + \tilde{a}_1^2 + \tilde{a}_2^2) - \frac{\tilde{a}_1 \tilde{a}_2 \beta}{4} - \omega(\delta^2 + \eta),\\
			&m_{3,3}=-2\alpha^2 + \beta + \frac{\beta}{192}(8\tilde{a}_0^4 + 12\tilde{a}_0^2\tilde{a}_1^2 + 36\tilde{a}_0^2 \tilde{a}_2^2 + \tilde{a}_1^4 + 6\tilde{a}_1^2 \tilde{a}_2^2 + 5\tilde{a}_2^4) - \beta\left(\frac{\tilde{a}_0^2}{2} + \frac{\tilde{a}_1^2}{8} + \frac{3\tilde{a}_2^2}{8}\right) - 1.
			\end{aligned}
		\end{equation}
Similarly, we get that the forcing term vector $\mathbf{F}_b(\mathbfit{a})$ appearing in equation~\eqref{E:M_b def} is given by
\begin{equation}\label{E:F_b}
	\mathbf{F}_b(\mathbfit{a})=\frac{1}{192}\begin{bmatrix}
		&0\\
		&192 (A\eta\omega^2 - \tilde{a}_2\beta\omega) - \tilde{a}_2\beta\omega(8\tilde{a}_0^4 + 12\tilde{a}_0^2\tilde{a}_1^2 + 12\tilde{a}_0^2 \tilde{a}_2^2 + \tilde{a}_1^4 + 2\tilde{a}_1^2 \tilde{a}_2^2 + \tilde{a}_2^4) + 4\tilde{a}_2\beta\omega(4\tilde{a}_0^2 + \tilde{a}_1^2 + \tilde{a}_2^2)\\
		&192\tilde{a}_1\beta\omega + \tilde{a}_1\beta\omega(8\tilde{a}_0^4 + 12\tilde{a}_0^2 \tilde{a}_1^2 + 12\tilde{a}_0^2 \tilde{a}_2^2 + \tilde{a}_1^4 + 2\tilde{a}_1^2 \tilde{a}_2^2 + \tilde{a}_2^4) - 24\tilde{a}_1\beta\omega(4\tilde{a}_0^2 + \tilde{a}_1^2 + \tilde{a}_2^2)
			\end{bmatrix}.
\end{equation}

Next, substituting the expansions given in~\eqref{E:harmonic balance} into equation~\eqref{E:basic system simplified expanded3}, it is possible to verify that the matrix $\mathbf{M}_c(\mathbfit{a})$ appearing in equation~\eqref{E:M_c def} is given by
\begin{equation}\label{E:M_c1}
	\mathbf{M}_c=\begin{bmatrix}
		&3 &0 &0 &0 &0\\
		&0 &-0.5\omega &-1.5 &0 &0\\
		&0 &-1.5 &0.5\omega &0 &0\\
		&0 &0 &0 &-\omega &-1.5\\
		&0 &0 &0 &-1.5 &\omega
	\end{bmatrix}.
\end{equation}
In addition, the forcing term vector $\mathbf{F}_c(\mathbfit{a},\mathbf{b})$ appearing in equation~\eqref{E:M_c def} is given by
{\fontsize{7}{10}\selectfont{
\begin{equation}\label{E:F_c}
	\mathbf{F}_c(\mathbfit{a},\mathbf{b})=\frac{1}{12}\begin{bmatrix}
		&8\tilde{a}_0^3 \tilde{b}_0 + 12\tilde{a}_0^2 \tilde{a}_1 \tilde{b}_1 + 12\tilde{a}_0^2 \tilde{a}_2 \tilde{b}_2 + 12\tilde{a}_0 12\tilde{a}_1^2 \tilde{b}_0 + 12\tilde{a}_0 \tilde{a}_2^2 \tilde{b}_0 - 12\tilde{a}_0 \tilde{b}_0 + 3\tilde{a}_1^3 \tilde{b}_1 + 3\tilde{a}_1^2 \tilde{a}_2 \tilde{b}_2+ 3\tilde{a}_1 \tilde{a}_2^2 \tilde{b}_1 - 6\tilde{a}_1 \tilde{b}_1 + 3\tilde{a}_2^3 \tilde{b}_2 - 6\tilde{a}_2 \tilde{b}_2 + 12\delta \tilde{b}_0^2 + 6\delta \tilde{b}_1^2 + 6\delta \tilde{b}_2^2
		\\
		&4\tilde{b}_2 \tilde{a}_0^3 + 12\tilde{b}_0 \tilde{a}_0^2 \tilde{a}_2 + 3\tilde{b}_2 \tilde{a}_0 \tilde{a}_1^2 + 6\tilde{b}_1 \tilde{a}_0 \tilde{a}_1 \tilde{a}_2 +9\tilde{b}_2 \tilde{a}_0 \tilde{a}_2^2 - 6\tilde{b}_2 \tilde{a}_0 + 3\tilde{b}_0 \tilde{a}_1^2 \tilde{a}_2 + 3\tilde{b}_0 \tilde{a}_2^3 - 6\tilde{b}_0 \tilde{a}_2 + 12\tilde{b}_0 \tilde{b}_2\delta\\
		&4\tilde{b}_1 \tilde{a}_0^3 + 12\tilde{b}_0 \tilde{a}_0^2 \tilde{a}_1 + 9\tilde{b}_1 \tilde{a}_0 \tilde{a}_1^2 + 6\tilde{b}_2 \tilde{a}_0 \tilde{a}_1 \tilde{a}_2 + 3\tilde{b}_1 \tilde{a}_0 \tilde{a}_2^2 - 6\tilde{b}_1 \tilde{a}_0 + 3\tilde{b}_0 \tilde{a}_1^3 + 3\tilde{b}_0 \tilde{a}_1 \tilde{a}_2^2 - 6\tilde{b}_0 \tilde{a}_1 + 12\tilde{b}_0 \tilde{b}_1\delta\\
		&6\tilde{b}_2 \tilde{a}_0^2 \tilde{a}_2 - 6\tilde{a}_1 \tilde{b}_1 \tilde{a}_0^2 + 6\tilde{b}_0 \tilde{a}_0 \tilde{a}_2^2 - 6\tilde{a}_1^2 \tilde{b}_0 \tilde{a}_0 + 2\tilde{b}_2 \tilde{a}_2^3 - 3\tilde{b}_2 \tilde{a}_2 + 3\tilde{a}_1 \tilde{b}_1 - 3\delta\left(\tilde{b}_1^2 - \tilde{b}_2^2\right) - 2\tilde{a}_1^3 \tilde{b}_1\\
		&6\tilde{b}_2 \tilde{a}_0^2 \tilde{a}_1 + 6\tilde{b}_1 \tilde{a}_0^2 \tilde{a}_2 + 12\tilde{b}_0 \tilde{a}_0 \tilde{a}_1 \tilde{a}_2 + \tilde{b}_2 \tilde{a}_1^3 + 3\tilde{b}_1 \tilde{a}_1^2 \tilde{a}_2 + 3\tilde{b}_2 \tilde{a}_1 \tilde{a}_2^2 - 3\tilde{b}_2 \tilde{a}_1 + \tilde{b}_1 \tilde{a}_2^3 - 3\tilde{b}_1 \tilde{a}_2 + 6\tilde{b}_1 \tilde{b}_2 \delta
			\end{bmatrix}.
	\end{equation}}}

In order to justify the expression for $\tilde{a}_0$ which was given in~\eqref{E:a0}, let us consider the equation for $\tilde{a}_0$ in the system~\eqref{E:harmonic balance nonlinear sys}, which follows from the harmonic balance of the equation in~\eqref{E:basic system simplified expanded1}. It is possible to verify that this equation is given by,	
\begin{equation}\label{E:gen form a0}
	\begin{aligned}
	F_{1}(\tilde{a}_0,\tilde{a}_1,\tilde{a}_2)&=\frac{\tilde{a}_0^3 \tilde{c}_0}{6} - \frac{\tilde{a}_1 \tilde{c}_1}{2} - \frac{\tilde{a}_2 \tilde{c}_2}{2} - \frac{\tilde{a}_0^2 \tilde{b}_0}{2} - \frac{\tilde{a}_1^2 \tilde{b}_0}{4} - \frac{\tilde{a}_2^2 \tilde{b}_0}{4} - \tilde{a}_0 \tilde{c}_0 + \frac{\tilde{a}_1^3 \tilde{c}_1}{16} + \frac{\tilde{a}_2^3 \tilde{c}_2}{16} - \tilde{b}_0(\beta - 1) - \frac{\tilde{a}_0 \tilde{a}_1 \tilde{b}_1}{2}\\
	& - \frac{\tilde{a}_0 \tilde{a}_2 \tilde{b}_2}{2} + \frac{\tilde{a}_0 \tilde{a}_1^2 \tilde{c}_0}{4} + \frac{\tilde{a}_0 \tilde{a}_2^2 \tilde{c}_0}{4} + \frac{\tilde{a}_0^2 \tilde{a}_1 \tilde{c}_1}{4} + \frac{\tilde{a}_1 \tilde{a}_2^2 \tilde{c}_1}{16} + \frac{\tilde{a}_0^2 \tilde{a}_2 \tilde{c}_2}{4} - \frac{\tilde{a}_0 \tilde{a}_1^2 \tilde{c}_4}{8} + \frac{\tilde{a}_1^2 \tilde{a}_2 \tilde{c}_2}{16} + \frac{\tilde{a}_0 \tilde{a}_2^2 \tilde{c}_4}{8} + \frac{\tilde{a}_0 \tilde{a}_1 \tilde{a}_2 \tilde{c}_3}{4}=0.
	\end{aligned}
\end{equation}
Now, using the definition of $\mathbf{b}$ and the expressions in~\eqref{E:M_b}-\eqref{E:F_b}, it is possible to verify that $\tilde{b}_0$, $\tilde{b}_1$, and $\tilde{b}_2$ take the form
\begin{equation}\label{E:b0}
	\begin{aligned}
		&\tilde{b}_0=\tilde{a}_0 g_0(\tilde{a}^2_0,\tilde{a}_1,\tilde{a}_2),\\
		&\tilde{b}_i=g_i(\tilde{a}^2_0,\tilde{a}_1,\tilde{a}_2),\quad i=1,2,
	\end{aligned}
\end{equation}
where $g_i(\tilde{a}^2_0,\tilde{a}_1,\tilde{a}_2)$, $i=0,1,2$, are rational functions of $\tilde{a}^2_0$, which are not singular at $\tilde{a}_0=0$. 
In order to prove that~\eqref{E:b0} holds, let us consider the inverse matrix $\mathbf{M}^{-1}_b(\mathbfit{a})=(\xi_{i,j})_{1\leq i,j\leq 3}$ and the forcing term vector $\mathbf{F}_b(\mathbfit{a})=(0,\varphi_{2}(\mathbfit{a}),\varphi_{3}(\mathbfit{a}))^T$, where  $\mathbf{M}_b(\mathbfit{a})$ and $\mathbf{F}_b(\mathbfit{a})$ are given explicitly in~\eqref{E:M_b}--\eqref{E:F_b}. Note that $b_i$, $i=0,1,2$, are obtained as follows
\begin{equation}\label{E:form b}
	b_i=(\xi_{i+1,2},\xi_{i+1,3})\cdot (\varphi_{2}(\mathbfit{a}),\varphi_{3}(\mathbfit{a})).
\end{equation}
Now, it is easy to verify that the elements of the matrix $\mathbf{M}^{-1}_b(\mathbfit{a})$ are rational functions of the form, 
\begin{equation}\label{E:form xi}
	\begin{aligned}
	&\xi_{1,j}=\tilde{a}_0\xi_{1,j}(\tilde{a}_0^2,\tilde{a}_1,\tilde{a}_2), \quad &&j=2,3,\\
	&\xi_{i,j}=\xi_{i,j}(\tilde{a}^2_0,\tilde{a}_1,\tilde{a}_2), \quad &&i,j=2,3,
	\end{aligned}
\end{equation}
which are not singular at $\tilde{a}_0=0$. Moreover, note that $\varphi_2(\mathbfit{a})$ and $\varphi_3(\mathbfit{a})$ are polynomials in $\tilde{a}^2_0$. Hence, from~\eqref{E:F_b}, \eqref{E:form b}, and~\eqref{E:form xi}, we may conclude that~\eqref{E:b0} holds.

Next, note that the matrix $\mathbf{M}_c$, which is given in~\eqref{E:M_c1}, is independent of $\tilde{a}_0$. Hence, looking on~\eqref{E:F_c} and using~\eqref{E:b0}, we may conclude that there exists rational functions $w_i$, $i=0,1,2,3,4$, of $\tilde{a}_0^2$, which are not singular at $\tilde{a}_0=0$, so that
\begin{equation}\label{E:form c_i}
	\begin{aligned}
		&\tilde{c}_i=w_i(\tilde{a}_0^2,\tilde{a}_1,\tilde{a}_2), &&i=0,3,4,\\
		&\tilde{c}_i=\tilde{a}_0 w_i(\tilde{a}_0^2,\tilde{a}_1,\tilde{a}_2), &&i=1,2.
	\end{aligned}
\end{equation}  
Finally, substituting~\eqref{E:b0} and~\eqref{E:form c_i} into~\eqref{E:gen form a0}, we may conclude that $\tilde{a}_0$ is of the form that was given in~\eqref{E:a0}.

\bigskip

For $A=0.1$, $\alpha=\beta=1/3$, $\eta=0.0118$, we get that the stability transition curve may be approximated by the following polynomial, which are rounded to fifth digit 
\begin{equation}\label{E:closed form pol coeffs}
		\begin{aligned}
			&d_8(\delta)\approx-9.0195\cdot10^{-3}\delta^7 + 4.7409\cdot10^{-4}\delta^5 + 1.4956\cdot10^{-5}\delta^3 + 9.5651\cdot10^{-7}\delta,\\
			&d_6(\delta)\approx- 1439.5\delta^8 - 3.5514\cdot10^{-2}\delta^7 - 67.943\delta^6 + 189.61\delta^5 - 1.2026\delta^4 + 4.48\delta^3\\
			&\phantom{d_6(\delta)\approx} - 9.4604\cdot10^{-3}\delta^2+2.6596\cdot10^{-2}\delta - 2.7908\cdot10^{-5},\\
			&d_4(\delta)\approx- 3238.8\delta^8 - 152.87\delta^6 + 2132.7\delta^5 - 2277.4\delta^4 + 50.353\delta^3 - 53.705\delta^2 + 150.08\delta - 0.3168,\\
			&d_2(\delta)\approx3838.61\delta^5 - 5118.1\delta^4 + 90.591\delta^3 - 120.79\delta^2 + 1685.52\delta - 899.37,\\
			&d_0(\delta)\approx3032.907\delta - 2021.98.
		\end{aligned}
\end{equation}

\end{document}